\documentclass{article}
\usepackage[eandd, final]{neurips_2026}

\usepackage[utf8]{inputenc}
\usepackage[T1]{fontenc}
\usepackage{hyperref}
\usepackage{url}
\usepackage{booktabs}
\usepackage{amsfonts}
\usepackage{nicefrac}
\usepackage{microtype}
\usepackage[table]{xcolor}
\usepackage{enumitem}
\usepackage{multirow}
\usepackage{graphicx}
\usepackage{makecell}
\usepackage{subcaption}
\usepackage{amsmath}
\usepackage{tabularx}
\usepackage{float}
\usepackage{placeins}
\usepackage{array}

\title{WorldArena 2.0: Extending Embodied World Model Benchmarking on Modality, Functionality and Platform}

\author{
Zhuohang Li$^{1,*}$~~~~~
Yiding Ma$^{1,*}$~~~~~
\textbf{Yu Shang}$^{1,*}$~~~~~
Lei Jin$^{1,*}$~~~~~
\textbf{Shengyuan Wang}$^{1}$\\
\textbf{Jianjie Fang}$^{1}$~~~~~
\textbf{Zhaolu Wang}$^{5}$~~~~~
\textbf{Xvyuan Liu}$^{1}$~~~~~
\textbf{Yinzhou Tang}$^{1}$\\
\textbf{Xin Jin}$^{1}$~~~~~
\textbf{Ziyou Wang}$^{1}$~~~~~
\textbf{Weikang Su}$^{1}$~~~~~
\textbf{Xin Zhang}$^{1}$~~~~~
\textbf{Haisheng Su}$^{2}$\\
\textbf{Weizhen He}$^{1}$~~~~~
\textbf{Wei Wu}$^{1}$~~~~~
\textbf{Haoyi Duan}$^{3}$~~~~~
\textbf{Gordon Wetzstein}$^{3}$~~~~~
\textbf{Xihui Liu}$^{6}$\\
\textbf{Dhruv Shah}$^{4}$~~~~~
\textbf{Zhaoxiang Zhang}$^{8}$~~~~~
\textbf{Zhibo Chen}$^{9}$~~~~~
\textbf{Jun Zhu}$^{1}$~~~~~
\textbf{Yonghong Tian}$^{7}$\\
\textbf{Tat-Seng Chua}$^{10}$~~~~~
\textbf{Xin Wang}$^{1}$~~~~~
\textbf{Wenwu Zhu}$^{1}$~~~~~
\textbf{Chen Gao}$^{1}$\footnotemark[2]~~~~~
\textbf{Yong Li}$^{1}$\footnotemark[2] \\
\\
$^{1}$Tsinghua University ~~
$^{2}$Shanghai Jiao Tong University ~~
$^{3}$Stanford University \\
$^{4}$Princeton University ~~
$^{5}$Zhongguancun Academy ~~
$^{6}$The University of Hong Kong \\
$^{7}$Peking University ~~
$^{8}$Chinese Academy of Sciences \\
$^{9}$University of Science and Technology of China \\
$^{10}$National University of Singapore ~~
}

\begin{document}

\footnotetext[1]{Equal contribution.}
\footnotetext[2]{Corresponding author: 
chgao96@gmail.com, liyong07@tsinghua.edu.cn}

\maketitle

\begin{abstract}World models have emerged as a central paradigm for embodied intelligence, enabling agents to predict action-conditioned future and reason about environmental dynamics. However, existing embodied world model benchmarks are still largely confined to vision-only prediction, offline embodied applications, and simulator-based evaluation, making them insufficient for assessing increasingly comprehensive world models. In this work, we introduce WorldArena 2.0, an expanded benchmark that systematically broadens embodied world model evaluation along three dimensions: \emph{modality}, \emph{functionality}, and \emph{platform}. Along the modality dimension, WorldArena 2.0 extends evaluation from vision-only to visuotactile modalities, enabling assessment of multimodal perception and prediction. Along the functionality dimension, it extends beyond policy evaluation and planning to assess world models as interactive RL environments for policy optimization. Along the platform dimension, it moves beyond simulator-only evaluation to a diverse suite of simulated and real-world robotic settings across multiple embodiments. Under a standardized protocol, WorldArena 2.0 comprehensively evaluates perceptual quality, interactive utility, and cross-platform performance, providing a comprehensive testbed for tracking progress toward embodied world models. The project page including codes and demos is available at: \hyperlink{https://world-arena.ai/}{https://world-arena.ai/}.

\end{abstract}

\section{Introduction}

World models~\citep{ding2025understanding,shang2026survey,long2025survey} are increasingly adopted as the core component for embodied intelligence, as they allow agents to predict future states and reason about the outcomes of their actions. In embodied applications, however, the practical value of a world model depends on more than visual realism. Beyond generating visually accurate future observations, these models need to capture physically grounded dynamics to support downstream reasoning and decision-making, such as action planning and policy learning. This transition from visual predictors to interactive environments naturally requires a corresponding shift in evaluation methods. As a result, modern benchmarks are gradually expanding their focus from visual quality to functional utility, aiming to assess these models under more realistic and demanding conditions.

Existing benchmarks for world models generally fall into two categories. The first one primarily focuses on the quality of video generation, evaluating models based on spatio-temporal consistency and visual fidelity~\citep{li2025worldmodelbench,duan2025worldscore,yue2025ewmbench}. The second category advances toward embodied functionality, assessing whether world models can capture action-conditioned dynamics to facilitate specific downstream tasks. Notable frameworks in this line include WorldSimBench~\citep{qin2024worldsimbench}, WorldEval~\citep{li2025worldeval}, World-in-World~\citep{zhang2025world}, WoW-World-Eval~\citep{fan2026wow} and WorldArena~\citep{shang2026worldarena}. 
Among these, WorldArena~\citep{shang2026worldarena} represents an important step toward unified embodied world model evaluation. It provides a systematic benchmark that jointly evaluates perceptual quality and downstream functional utility, moving beyond purely visual prediction to examine whether world models can support embodied decision-making. In particular, WorldArena introduces a standardized evaluation framework in which world models are assessed not only by the fidelity of generated videos, but also by their usefulness for embodied tasks, including data engine, action planning and policy evaluation. This design establishes an explicit bridge between generative prediction and end-to-end system utility. Rather than treating world models as isolated video predictors, our benchmark assesses their practical effectiveness when integrated with downstream control interfaces.

While these benchmarks provide a foundation for evaluating visual quality and basic task functionality, they do not fully capture the interactive and physical requirements of actual embodied deployment. Specifically, there remain three major limitations. First, prevailing benchmarks are predominantly confined to vision-only settings. This overlooks the multimodal nature of embodied interactions, where tactile feedback is essential for resolving contact-rich dynamics and physical friction. Second, downstream evaluation is largely restricted to open-loop planning~\citep{zhang2025world} or static policy evaluation~\citep{li2025worldeval}, rarely investigating the capacity of a world model to serve as an interactive reinforcement learning environment that supports iterative policy improvement through imagined rollouts. Finally, benchmark results are currently obtained almost exclusively in simulation, leaving it entirely unclear whether strong performance in controlled virtual environments translates to real-world deployment.

To address these limitations, we introduce \textbf{WorldArena 2.0}, which extends WorldArena along three coordinated dimensions: \emph{modality}, \emph{functionality}, and \emph{platform}.
First, WorldArena 2.0 expands benchmarking from visual to visuotactile world models. By developing a standardized framework built upon the UniVTAC simulator~\citep{chen2026univtac}, we enable visual world models to perceive and predict richer, multimodal sensory streams. This extension is critical for embodied manipulation, where tactile signals provide direct evidence of contact, force, slip, and material interaction that cannot be reliably inferred from visual observations alone. By incorporating simulated tactile prediction and a matched real-robot tactile ablation into benchmark evaluation, WorldArena 2.0 evaluates multimodal prediction alongside physical manipulation.
Second, WorldArena 2.0 broadens functional evaluation by using world models as interactive RL training environments. Rather than merely evaluating frozen policies, our framework uses learned environments for policy optimization. This setting provides a more demanding test of world model utility: the model must remain stable and action-consistent over iterative interactions, produce reward-relevant state transitions, and support policy optimization without severe compounding errors. With support for diverse action-conditioned world models and modular reward designs, WorldArena 2.0 evaluates whether training in learned dynamics improves policy performance in the ground-truth simulator, rather than merely predicting plausible futures.
Finally, along the platform dimension, WorldArena 2.0 moves beyond simulator-centric evaluation to a comprehensive suite of simulated and real-world testbeds. In addition to two multi-embodiment simulation environments, RoboTwin 2.0~\citep{chen2025robotwin} and LIBERO~\citep{liu2023libero}, we introduce physical robot evaluation on the AgileX Split-Type ALOHA platform. We benchmark physical manipulation across distinct operational protocols: pouring water and table wiping for cross-platform evaluation, chip grasping and table wiping for physical tactile ablation (Appendix~\ref{sec:tactile_physical}), and a pilot investigation on bottle lifting for policy transfer (Appendix~\ref{sec:additional_diagnostics}). Together, these extensions establish WorldArena 2.0 as a rigorous diagnostic benchmark capable of assessing world models under grounded perception and physical interaction constraints. Our main contributions are threefold:
\begin{itemize}[leftmargin=*]
    \item We introduce WorldArena 2.0, an expanded diagnostic benchmark systematically evaluating embodied world models across three dimensions: modality, functionality, and platform.
    \item We establish standardized evaluation protocols that incorporate visuotactile dynamics and evaluate world models as interactive RL training environments for policy optimization.
    \item We conduct extensive evaluations across simulated and physical robotic platforms, revealing that while learned models yield tangible policy-optimization gains in simulation, severe reality gaps remain in sustained contact-rich manipulation and physical transfer.
\end{itemize}

\begin{figure*}[t]
    \centering
    \includegraphics[width=0.9\textwidth]{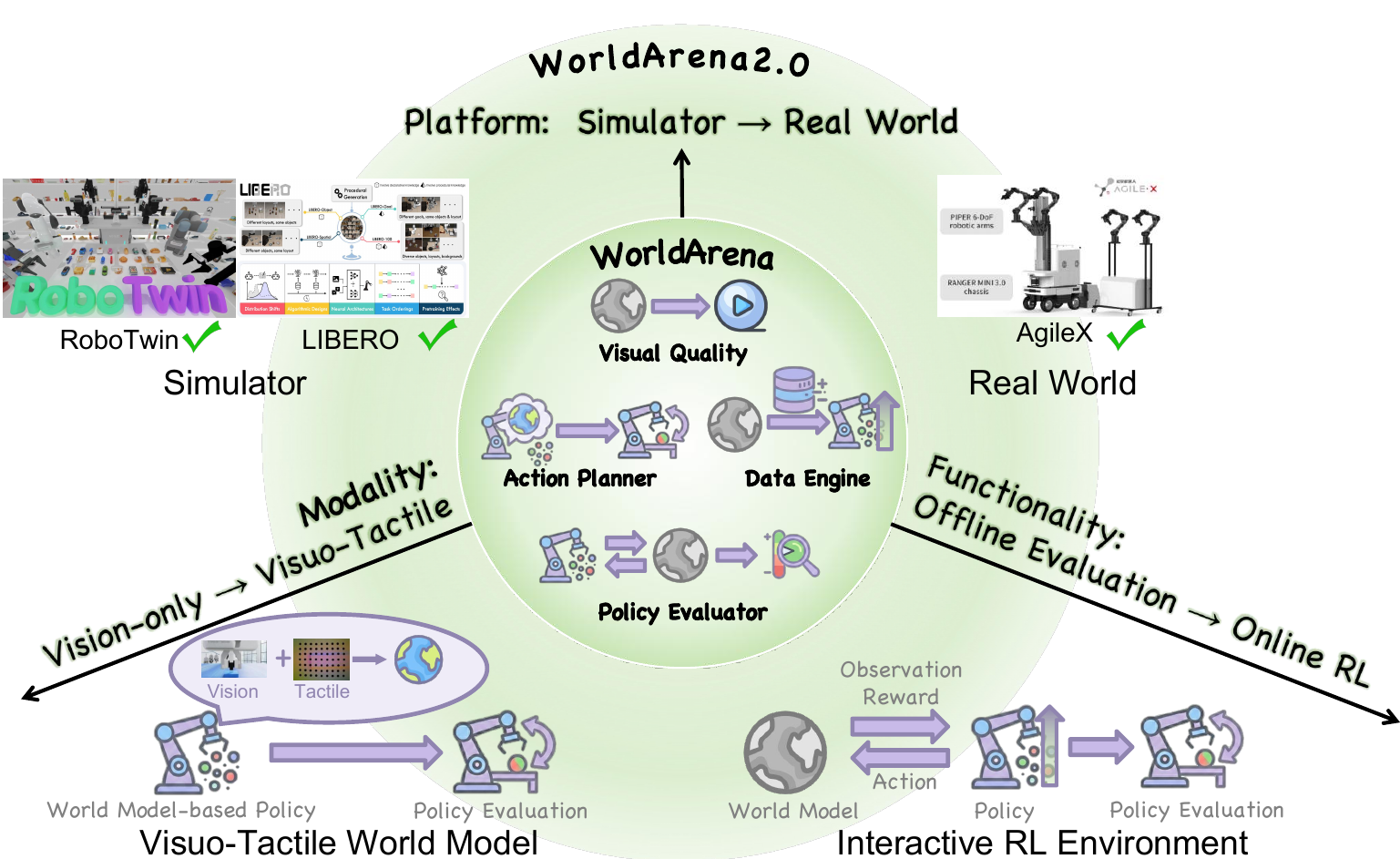}
\caption{Overview of the extension from WorldArena to WorldArena 2.0 along three dimensions: modality, functionality, and platform.}
\vspace{-0.5cm}
    \label{fig:combined}
\end{figure*}


\section{Related Work}

\subsection{Embodied World Models}
Embodied world models (EWMs) aim to capture environment dynamics to support decision-making, planning, and counterfactual reasoning in embodied settings~\citep{shang2026survey,long2025survey,li2025comprehensive}. Existing approaches fall into three broad categories. The first learns compact latent dynamics for control and reinforcement learning, exemplified by the Dreamer family~\citep{hafnermastering,hafner2025mastering}, which improves sample efficiency in simulation. The second leverages large-scale generative models~\citep{wan2025wan,yang2024cogvideox,zhu2024sora} to perform action-conditioned video prediction~\citep{gu2025cosmos,shang2025roboscape,guo2025ctrl,chen2026abot}, generating future visual observations conditioned on agent actions and language instructions via diffusion~\citep{chi2025wow,feng2025vidar,zhu2024irasim} or autoregressive architectures~\citep{wu2024ivideogpt}. The third focuses on enhancing physical and interaction fidelity through physics-informed objectives and joint visual-action modeling~\citep{bi2025motus,ye2026world,li2026causal,yuan2026fast}. Together, these directions reflect a shift from low-dimensional dynamics to visually grounded, action-aware prediction.

Despite these advances, visual realism does not guarantee physical validity. Generative rollouts frequently violate basic physical constraints and suffer from error accumulation over extended horizons. This discrepancy demands evaluation paradigms that move beyond visual quality to assess multimodal dynamics grounding and closed-loop interactive decision-making across simulated and physical platforms.

\subsection{Benchmarks for Embodied World Models}
As generative models are increasingly applied in embodied domains, evaluating their effectiveness as world models remains challenging. Standard video generation metrics~\citep{duan2025worldscore,lu20254dworldbench} emphasize perceptual quality but largely overlook physical realism and action-relevant fidelity. Recent benchmarks can be grouped into three types. The first type focuses on perceptual and spatiotemporal quality, e.g., EWMBench~\citep{yue2025ewmbench}, which measures scene consistency, motion correctness, and semantic alignment. While informative for visual fidelity, such benchmarks do not assess whether generated dynamics support embodied decision-making. The second type evaluates functional utility, such as WorldSimBench~\citep{qin2024worldsimbench}, which translates generated videos into control signals via inverse dynamics, or WorldEval~\citep{li2025worldeval} and WoW-World-Eval~\citep{fan2026wow}, which test policy evaluation or action planning. The third type provides unified frameworks, exemplified by WorldArena~\citep{shang2026worldarena}, which jointly measures perceptual quality and multiple functional aspects, including data engine, policy evaluation, and action planning.

Despite these advances, existing benchmarks have important limitations. They rely exclusively on visual inputs, ignoring tactile feedback that is critical for real-world physical interactions. Their functional evaluations are largely restricted to fixed policies or single-action planning and rarely test whether a world model can support continuous, interactive policy training without error accumulation. Furthermore, most assessments remain simulator-only, leaving the sim-to-real gap largely unexamined. WorldArena 2.0 addresses these gaps by incorporating visuotactile modalities, supporting policy optimization in learned environments, and evaluating performance across both simulated and real robotic platforms, bringing benchmark evaluation closer to real-world embodied requirements.

\section{WorldArena 2.0 Benchmark}

\subsection{From WorldArena to WorldArena 2.0}
WorldArena~\citep{shang2026worldarena} pioneered the joint evaluation of perceptual fidelity (16 metrics across 6 dimensions) and offline downstream utility (data engine, policy evaluation, and action planning). However, its scope was restricted to vision-only simulation. WorldArena 2.0 expands this foundation along three coordinated dimensions: (1)~\textbf{Modality}, incorporating visuotactile dynamics to resolve contact-rich physical interactions; (2)~\textbf{Functionality}, evaluating world models as interactive RL training environments that support iterative policy optimization; and (3)~\textbf{Platform}, bridging multi-embodiment simulation (RoboTwin 2.0, LIBERO) and physical execution on an AgileX dual-arm robot. This unified design transforms embodied world model benchmarking from passive video verification into an interactive, physically grounded diagnostic testbed.

\begin{figure*}[t]
    \centering
    \includegraphics[width=\textwidth]{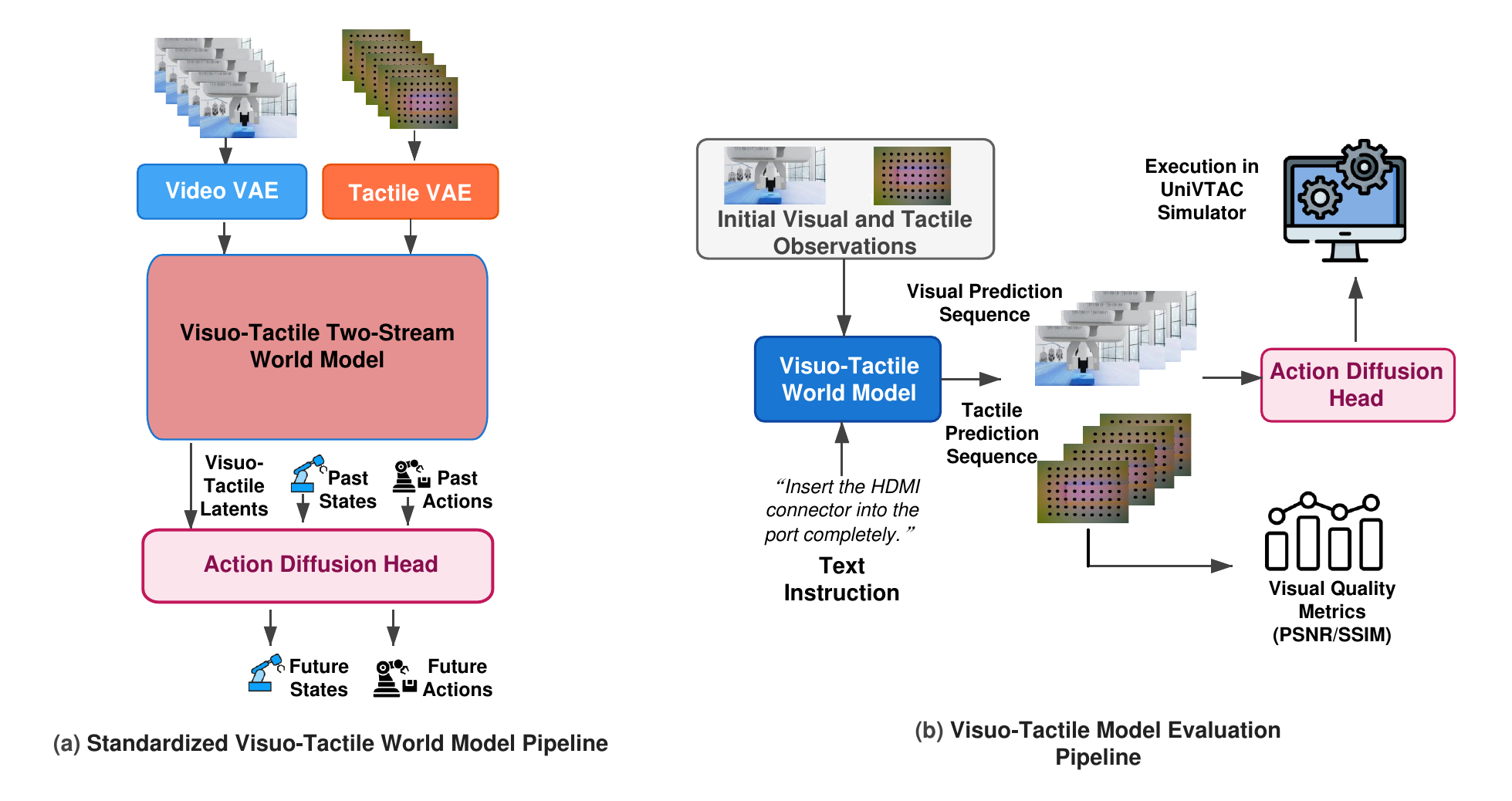}
    \caption{The standardized visuotactile world model architecture design (a) and the visuotactile model evaluation pipeline based on the UniVTAC simulator (b).}
    \label{fig:vtwm}
\end{figure*}

\subsection{Modality Extension: Evaluating Visuotactile World Models}
While contemporary world models~\citep{chi2025wow,feng2025vidar,liao2025genie,team2025gigaworld} predominantly focus on visual perception, vision-only paradigms remain fundamentally insufficient for contact-rich manipulation tasks, where critical interaction dynamics are only partially observable through visual cues, and diverge substantially from human multimodal sensory experience.
Recent efforts have constructed visuotactile world models through joint visuotactile modeling and tactile encoding injection, validating the effectiveness of incorporating tactile information in contact-rich scenarios~\citep{yuan2026vtam,higuera2026VTWM,zheng2026omnivta}. However, these works operate under heterogeneous setups with divergent tactile sensor configurations and inconsistent evaluation protocols.
Consequently, the field still lacks a standardized evaluation pipeline for visuotactile world models, and existing benchmarks~\citep{fan2026wow,huang2024vbench} provide no coverage of tactile modalities. 
To evaluate the capabilities of existing vision-based embodied world models under multimodal perceptual conditions, WorldArena extends the evaluation scope from vision-only settings to visuotactile world models.

To address the aforementioned challenges, we construct a standardized pipeline that upgrades vision-only world models to visuotactile world models.
As illustrated in Figure~\ref{fig:vtwm}, our pipeline builds upon an existing video world model, augmenting it with three primary modules: a tactile VAE, a visuotactile two-stream world model, and an action diffusion head.
First, the tactile VAE encodes sequences of tactile deformation maps and aligns them into the video latent space of the original world model, enabling plug-in augmentation without architectural surgery.
Second, the visuotactile two-stream world model performs synchronized denoising for both video prediction and tactile perception prediction, preserving modality-specific dynamics while enabling cross-modal coordination.
Finally, for downstream task execution, the action diffusion head receives past states and actions, together with the predicted visuotactile latents, and performs denoising to directly infer future actions, closing the loop between perceptual prediction and functional manipulation.
This standardized pipeline follows a modular design philosophy, augmenting existing video world models~\citep{chi2025wow,feng2025vidar,wan2025wan,liao2025genie} with plug-in tactile components rather than altering their underlying architectures.

Since the action decoder is trained separately for each world model, downstream task success measures the utility of the complete world-model--decoder system (Appendix~\ref{sec:interfaces}). Crucially, our future-information ablation confirms that downstream manipulation is actively driven by predictive dynamics rather than history memorization: conditioning on ground-truth future observations yields 66\% task success on Adjust Bottle, whereas removing future inputs (32\%) or temporally shuffling future frames (33\%) causes an immediate 50\% performance drop (Appendix~\ref{sec:future_ablation}). This verifies that the action decoder genuinely queries future predictive cues, ensuring that control performance reflects world-model dynamics fidelity.
This extension is based on the UniVTAC simulator ~\citep{chen2026univtac}. 
We leverage the official dataset provided by UniVTAC and construct a standardized evaluation pipeline for visuotactile world models within the simulated environment.
These extended models are assessed across two complementary dimensions: (1) perceptual quality of tactile video prediction, and (2) functional success rate in downstream manipulation tasks.
Through these standardized extensions, we broaden WorldArena's evaluated modality from vision-only perception to the visuotactile domain.

\subsection{Functionality Extension: Evaluating World Models as RL Environments}
\begin{figure*}[t]
    \centering
    \includegraphics[width=0.95\textwidth]{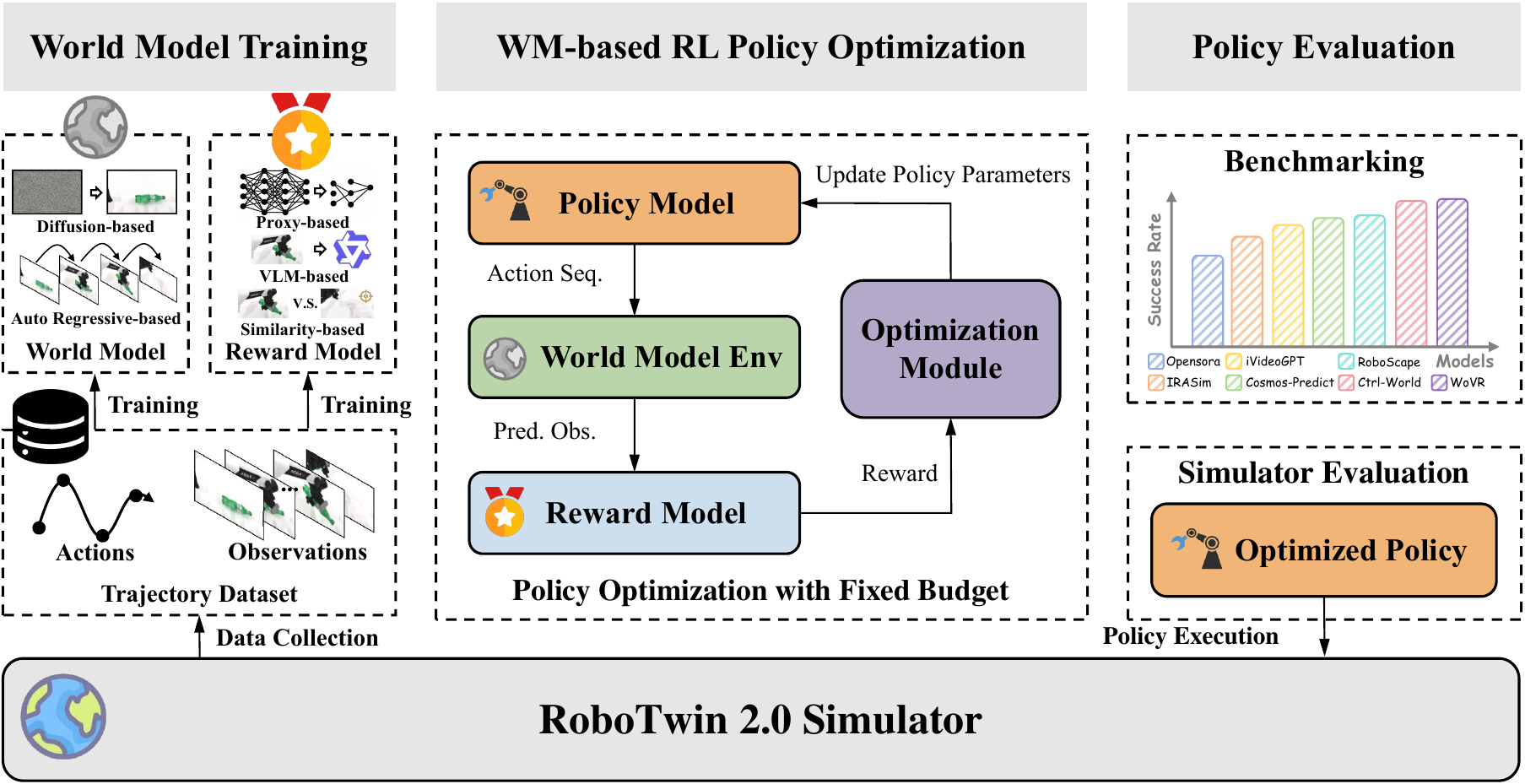}
\caption{WorldArena 2.0 framework for leveraging world models as RL environments, comprising three stages: world model training, RL policy optimization using the world model, and policy evaluation.}
    \label{fig:wm4rl}
\end{figure*}

World models have demonstrated strong capability in capturing environment dynamics from real-world data, but their evaluation has mostly been limited to offline prediction. As their fidelity improves, they can serve as interactive proxies for real environments, enabling reinforcement learning (RL) directly within the learned model. Recent studies have begun exploring the use of world models as virtual environments to support RL training~\citep{xiao2025world,tang2025roboscape,zhu2025wmpo,wu2024ivideogpt}. However, existing approaches often rely on system-level designs without standardized interfaces, making it difficult to isolate the contributions of individual components and compare their impact on final performance.

We propose a standardized framework (Figure~\ref{fig:wm4rl}) with three stages: world model training, policy optimization in the learned environment, and policy evaluation in the ground-truth simulator. The world model $\hat{P}_\phi$ predicts future observations, the policy $\pi_\theta$ selects actions, and the reward model $\hat{R}_\psi$ provides training signals. Let $h_t$ denote the available interaction history, including the task instruction. Their recursive interaction is
\begin{equation}
 a_t\sim\pi_\theta(\cdot\mid h_t),\quad
 \hat{o}_{t+1}\sim\hat{P}_\phi(\cdot\mid h_t,a_t),\quad
 \hat r_t=\hat R_\psi(h_t,a_t,\hat o_{t+1}).
 \label{eq:wm_recursion}
\end{equation}
Each component uses the context supported by its interface; this notation does not assume that a single observation is a Markov state. The partial-observability formulation and component-specific inputs are detailed in Appendix~\ref{sec:formalism}.

We first adapt the world and reward models using simulator trajectories. With these models fixed, actor-only GRPO~\citep{shao2024deepseekmath} updates the policy from group-relative advantages computed over rollouts collected in each learned environment. Finally, optimized policies execute in RoboTwin 2.0 under ground-truth dynamics, and its task-success predicate determines the reported success rates. This measures the utility of a world model as an RL training environment; physical transfer is evaluated separately. Implementation and evaluation protocols are provided in Appendix~\ref{sec:wm4rl_app}.

\subsection{Platform Extension: Cross-Embodiment Sim-to-Real Evaluation}

\begin{figure*}[t]
    \centering
    \includegraphics[width=\textwidth]{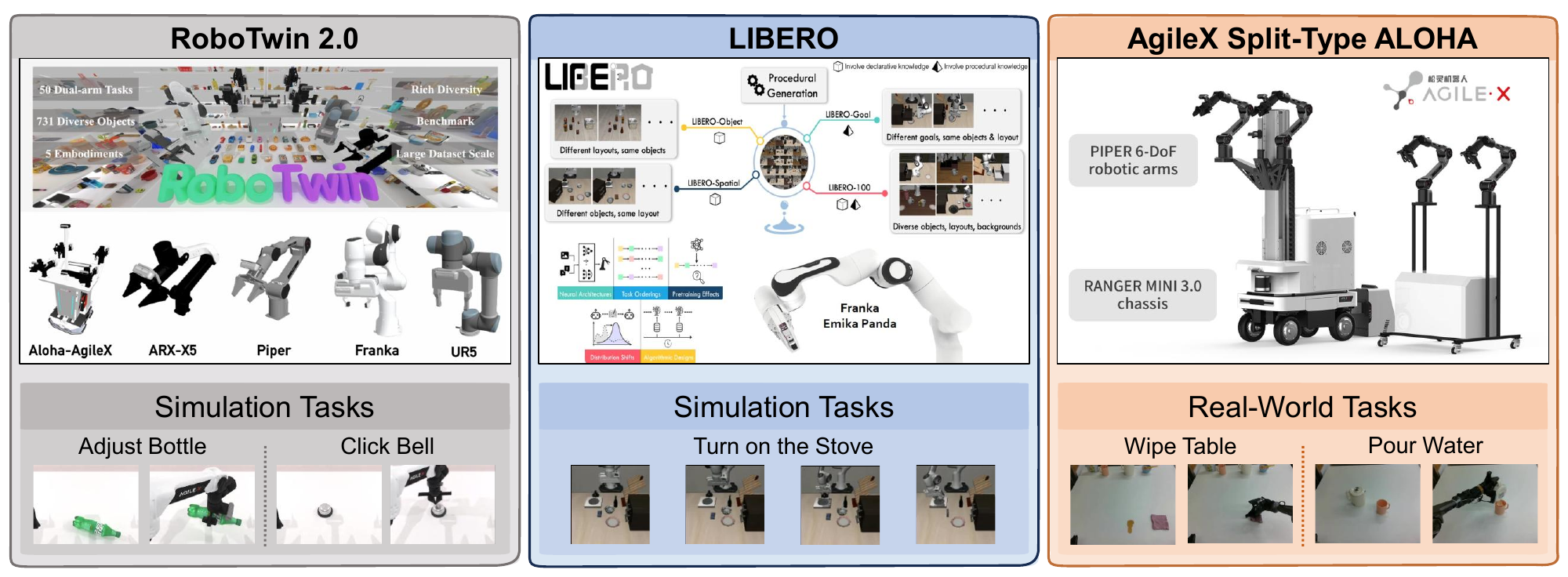}
    \caption{Overview of three test platforms in WorldArena 2.0: RoboTwin 2.0, LIBERO, and a real-world AgileX Split-Type ALOHA robotic platform.}
    \label{fig:platform}
    \vspace{-0.5cm}
\end{figure*}

Most existing benchmarks~\citep{yue2025ewmbench,li2025worldeval} for embodied world models predominantly evaluate prediction quality on a single simulator with a fixed task set and limited object variation. While this narrow scope simplifies evaluation, it makes results susceptible to overfitting, leading to artificially inflated rankings. Furthermore, conclusions drawn from a single platform are unstable by construction, as they provide no signal on whether a model's learned visual and physical priors can withstand the distribution shifts encountered on different robotic morphologies or real-world sensing conditions. To overcome these limitations, WorldArena 2.0 introduces a cross-embodiment evaluation protocol spanning three distinct platforms, RoboTwin 2.0, LIBERO, and an AgileX split-body ALOHA robot, forming complementary evaluation regimes. Rather than forcing identical, synthetic object configurations across all embodiments, we benchmark on each platform's representative native task distribution (Figure~\ref{fig:platform}). This assesses whether a world model's physical and visual priors maintain consistent functional utility across diverse dynamics regimes rather than overfitting to a single task morphology.

To further assess the functional utility of world models across these platforms, we adopt the two evaluation protocols introduced in WorldArena: (1) \textbf{Embodied Data Engine}, where synthetic trajectories generated by a world model are used to train downstream policies, with success rate on the target task measuring the data quality; (2) \textbf{Embodied Action Planner}, where an action decoder converts world-model futures into action chunks for repeated planning, and task completion rate reflects its planning reliability. By running these protocols across RoboTwin, LIBERO, and the AgileX ALOHA platform, we systematically measure whether improvements in prediction fidelity translate into real-robot task competence.

\section{Experiments}

\subsection{Visuotactile World Model Evaluation}
We conduct a unified evaluation on the UniVTAC simulator across two contact-rich manipulation tasks, \textit{Insert HDMI} and \textit{Lift Bottle}, assessing vision-based embodied world models that have been extended through standardized tactile injection and jointly trained.
We select representative vision-based embodied world models, including Vidar~\citep{feng2025vidar}, Wan2.2~\citep{wan2025wan}, Genie Envisioner~\citep{liao2025genie}, and ACT~\citep{zhao2023ACT} with a tactile tokenizer ~\citep{chen2026univtac}, as our baseline. 
For tactile prediction quality, we adopt PSNR and SSIM~\citep{wang2004ssim} as evaluation metrics; for downstream action execution, we measure task success rate. 
Tactile prediction results and downstream task success rates are summarized in Table~\ref{tab:tactile_results}.

Wan 2.2~\citep{wan2025wan} achieves the highest tactile prediction scores and reaches 100\% success on Insert HDMI, outperforming Vidar (70\%) and Genie Envisioner (0\%). However, all world-model-based policies fail on Lift Bottle, where the ACT baseline attains 80\%. Notably, while Genie Envisioner achieves higher tactile SSIM (0.456) than Vidar (0.278), it yields 0\% task success compared to Vidar's 70\%. This empirical discrepancy demonstrates that standard frame-level reconstruction metrics (PSNR/SSIM) do not monotonically correlate with downstream control. Rather, success hinges on whether the model accurately preserves fine-grained physical dynamics (e.g., contact timing and force transitions), reinforcing the necessity of evaluating closed-loop functional utility alongside perceptual quality.

\textbf{Real tactile evaluation.} Augmenting FastWAM~\citep{yuan2026fast} with tactile sensing increases average physical success from 17.5\% to 37.5\% across Chip (Grasp chip) and Wiping (Wipe table), with the most pronounced gain on the contact-sensitive Chip task (15\% $\rightarrow$ 45\%; Table~\ref{tab:real_tactile_added}). These physical findings complement our simulation diagnosis, providing direct evidence that tactile feedback is indispensable for contact-rich manipulation. Full trial counts, Wilson confidence intervals, and experimental protocols are detailed in Appendices~\ref{sec:tactile_physical} and~\ref{sec:trial_uncertainty}.

\begin{table}[t]
\centering
\caption{Visuotactile prediction and task success on UniVTAC. Trial counts and uncertainty are reported in Appendix~\ref{sec:trial_uncertainty}.}
\label{tab:tactile_results}

\fontsize{8}{9}\selectfont 

\setlength{\tabcolsep}{8pt} 

\setlength{\arrayrulewidth}{0.5pt}

\begin{tabular}{lcc|cc|c}
\toprule
\multirow{2}{*}{\textbf{Model}} & \multicolumn{2}{c|}{\textbf{Tactile Prediction Quality}} & \multicolumn{3}{c}{\textbf{Task Success Rate (\%)}} \\
\cmidrule(lr){2-3} \cmidrule(lr){4-6}
& \textbf{PSNR} $\uparrow$ & \textbf{SSIM} $\uparrow$ & \textbf{Insert HDMI} & \textbf{Lift Bottle} & \textbf{Avg.} \\
\midrule
ACT (Baseline) & -- & -- & 20 & \textbf{80} & \textbf{50} \\
Vidar          & 13.97 & 0.278 & 70 & 0 & 35 \\
Genie Envisioner  & 13.36 & 0.456 & 0 & 0 & 0 \\
Wan2.2   & \textbf{21.26} & \textbf{0.746} & \textbf{100} & 0 & \textbf{50} \\
\bottomrule
\end{tabular}

\end{table}


\subsection{World Model Evaluation as RL Environments}
We evaluate policies trained in seven world-model environments on two RoboTwin 2.0 tasks under three reward designs. Implementation and evaluation details are provided in Appendix~\ref{sec:wm4rl_app}.
\begin{table*}[t]
\centering
\small
\setlength{\tabcolsep}{6pt}
\renewcommand{\arraystretch}{1.08}

\caption{
Policy success rates (\%) on RoboTwin 2.0.
\textbf{Panel A} reports common baselines: SFT uses no RL reward, while
Simulator-based RL is trained with the native RoboTwin task reward.
\textbf{Panel B} compares policies optimized in learned world-model environments
under 3 reward designs.
All policies are evaluated using the same RoboTwin task-success criterion.
}
\label{tab:wm_rl}

\begin{tabular}{l l cc}
\toprule
\multicolumn{4}{l}{\textbf{(A) Common baselines}} \\
\midrule
\textbf{Method} &
\textbf{Training Reward} &
\begin{tabular}[c]{@{}c@{}}\textbf{Click}\\\textbf{Bell}\end{tabular} &
\begin{tabular}[c]{@{}c@{}}\textbf{Adjust}\\\textbf{Bottle}\end{tabular} \\
\midrule
SFT &
-- &
43.75 &
55.08 \\

Simulator-based RL &
Native reward &
87.30 &
78.90 \\
\bottomrule
\end{tabular}

\vspace{0.8em}

\begin{tabular}{lcccccc}
\toprule
\multicolumn{7}{l}{\textbf{(B) RL policy optimization in interactive world-model environments}} \\
\midrule

\multirow{2}{*}{\textbf{World Model}} &
\multicolumn{2}{c}{\textbf{Proxy-based}} &
\multicolumn{2}{c}{\textbf{VLM-based}} &
\multicolumn{2}{c}{\textbf{Similarity-based}} \\
\cmidrule(lr){2-3}
\cmidrule(lr){4-5}
\cmidrule(lr){6-7}

&
\begin{tabular}[c]{@{}c@{}}Click\\Bell\end{tabular} &
\begin{tabular}[c]{@{}c@{}}Adjust\\Bottle\end{tabular} &
\begin{tabular}[c]{@{}c@{}}Click\\Bell\end{tabular} &
\begin{tabular}[c]{@{}c@{}}Adjust\\Bottle\end{tabular} &
\begin{tabular}[c]{@{}c@{}}Click\\Bell\end{tabular} &
\begin{tabular}[c]{@{}c@{}}Adjust\\Bottle\end{tabular} \\
\midrule

OpenSora~\citep{zheng2024open}
& 56.25 & 60.16
& 55.27 & 57.03
& 53.13 & 58.00 \\

IRASim
& 53.13 & 61.33
& 53.52 & 58.98
& 50.78 & 59.38 \\

iVideoGPT
& 52.53 & 56.25
& 48.44 & 58.59
& 52.15 & 60.93 \\

Cosmos-Predict-2.5(action)~\citep{nvidia2025cosmospredict25}
& 67.38 & 63.48
& 54.10 & 58.40
& 63.09 & 61.13 \\

RoboScape
& 68.75 & 60.74
& 55.46 & 59.38
& 63.48 & 59.18 \\

Ctrl-World
& 69.53 & \textbf{70.70}
& 66.80 & \textbf{65.04}
& 69.92 & \textbf{66.02} \\

WoVR
& \textbf{75.00} & 67.19
& \textbf{69.38} & 64.45
& \textbf{72.07} & 61.35 \\

\bottomrule
\end{tabular}

\end{table*}

\textbf{Overall performance.}
Policy optimization inside learned world-model environments consistently improves downstream simulator success over the supervised fine-tuning (SFT) baseline across all tested configurations, although a gap remains relative to policies trained directly in the native simulator (Table~\ref{tab:wm_rl}). To assess the maximal policy-improvement capacity supported by each learned simulator, reported simulator scores represent the optimization ceiling across the budget, evaluated on held-out RoboTwin validation configurations (Appendix~\ref{sec:statistical_protocol}). WoVR~\citep{jiang2026wovr} leads the learned environments on Click Bell, whereas Ctrl-World~\citep{guo2025ctrl} leads on Adjust Bottle, reaching 75.00\% and 70.70\% under proxy rewards compared to 43.75\% and 55.08\% for SFT. For the same RL-environment role, DreamX-Phi~\citep{team2026dreamx} reaches 67.19\% (Adjust Bottle task and proxy-based reward) success for a separate policy trained in its learned environment, which numerically matches WoVR and lies between Cosmos-Predict-2.5(action) (63.48\%) and Ctrl-World (70.70\%) in Table~\ref{tab:wm_rl}. Given the substantial computational cost of iterative rollouts (profiling up to 28.8 minutes per iteration on $2\times$A800 GPUs; Appendix~\ref{sec:compute}), this comprehensive 42-run study operates under a standardized protocol with fixed iteration budgets. Notably, model rankings remain positively concordant across all three reward formulations (Spearman $\rho \in [0.706, 0.957]$ strictly within the seven learned models, Table~\ref{tab:wm_only_correlations}; and $\rho \ge 0.845$ when pooled across all baselines, Table~\ref{tab:reward_correlations}), verifying that relative model utility remains robust across diverse reward specifications.

\textbf{Reward-model design.}
We compare a learned visual proxy, a zero-shot VLM evaluator, and goal-image similarity rewards. WoVR and Ctrl-World retain their respective task-leading positions across all three reward families. This agreement supports the leading-model comparison across reward choices, while detailed correlations and remaining reward--generator dependencies are analyzed in Appendix~\ref{sec:reward_correlation}.

\textbf{Physical transfer.} On the real AgileX robot (Lift Bottle), policies optimized inside WoVR and Ctrl-World achieve 25\% and 30\% multi-stage task progress, substantially outperforming the SFT baseline (10\%; Appendix~\ref{sec:additional_diagnostics}). Extended analyses on rollout horizons, reward stability, and computational budgets are provided in Appendices~\ref{sec:transfer_diagnostics} and~\ref{sec:compute}.


\subsection{Cross-platform Evaluation and Analysis}
\label{sec:cross-plat}
\textbf{Cross-platform video quality evaluation.}
Following the evaluation protocol established in WorldArena~\citep{shang2026worldarena}, we evaluate video quality across six perceptual dimensions with 16 standardized metrics on RoboTwin, LIBERO, and the real-world AgileX Split-Type ALOHA platform. Detailed results are reported in Tables~\ref{tab:vq_robotwin_part1}, \ref{tab:vq_robotwin_part2}, \ref{tab:vq_libero_part1}, \ref{tab:vq_libero_part2}, \ref{tab:vq_real_part1}, \ref{tab:vq_real_part2}. A consistent pattern emerges across platforms. Commercial generative models such as Veo 3.1 and Wan 2.6 achieve state-of-the-art visual quality, yet they exhibit severe degradation on physical dynamics. On trajectory accuracy (Table~\ref{tab:vq_robotwin_part2}), specialized embodied world models like CtrlWorld and IRASim outperform Veo 3.1 and Wan 2.6 by more than $3\times$. This stark divergence highlights that high aesthetic and semantic fidelity does not guarantee physically grounded dynamics, underscoring the critical need for embodied evaluation. Among embodied models, WoW and CtrlWorld consistently rank highly on physics adherence and content consistency.

\textbf{Cross-platform task success rate.}
Table~\ref{tab:engine_planner} evaluates world models as data engines and action planners~\citep{shang2026worldarena} on RoboTwin, LIBERO, and the real-world AgileX platform. Several architectures achieve near-zero physical success, exposing a substantial gap relative to simulator performance. The extended real-robot suite likewise reveals uneven task competence, with precision-demanding cardboard-box folding remaining more difficult than wiping or garment folding (Appendix~\ref{sec:community_real_extension}). These observations motivate direct evaluation of physical task completion rather than inferring deployment capability from visual quality or simulator performance alone.

\begin{table}[t]
\centering
\caption{Task success rates (\%) of world models as embodied data engines and action planners across three platforms. Real-world Task 1/2 are Pour Water/Wipe Table. Trial counts and uncertainty are reported in Appendix~\ref{sec:platform_uncertainty}.}
\label{tab:engine_planner}
\small
\resizebox{\textwidth}{!}{
\begin{tabular}{l c c c c c c c c c c}
\toprule
 & \multicolumn{4}{c}{\textbf{RoboTwin 2.0}} & \multicolumn{2}{c}{\textbf{LIBERO}} & \multicolumn{4}{c}{\textbf{Real-World}} \\
\cmidrule(lr){2-5} \cmidrule(lr){6-7} \cmidrule(lr){8-11}
\multirow{2}{*}{\textbf{Model}} 
 & \multicolumn{2}{c}{\textbf{Data Engine}} & \multicolumn{2}{c}{\textbf{Action Planner}} 
 & \multicolumn{1}{c}{\textbf{Data}} & \multicolumn{1}{c}{\textbf{Action}}
 & \multicolumn{2}{c}{\textbf{Data Engine}} & \multicolumn{2}{c}{\textbf{Action Planner}} \\
\cmidrule(lr){2-3} \cmidrule(lr){4-5}
\cmidrule(lr){8-9} \cmidrule(lr){10-11}
 & Task~1 & Task~2 & Task~1 & Task~2 
 & \textbf{Engine} & \textbf{Planner}
 & Task~1 & Task~2 & Task~1 & Task~2 \\
\midrule
GigaWorld           & 2 & 13 & 6 & 19 & 0 & 0 & 0 & 0 & 0 & 0 \\
Genie Envisioner    & 7 & 21 & 10 & 20 & 2 & 6 & 0 & 0 & 0 & 20 \\
TesserAct~\citep{zhen2025tesseract}           & 1 & 35 & 1 & 35 & 34 & 38 & 0 & 0 & 0 & 30 \\
Vidar               & 13 & 53 & 2 & 19 & 22 & 14 & 40 & 0 & 30 & 10 \\
Wan~2.2             & 15 & 41 & 12 & 20 & 10 & 24 & 10 & 0 & 10 & 0 \\
CogVideoX           & 3 & 28 & 8 & 16 & 0 & 2 & 10 & 10 & 0 & 50 \\
\bottomrule
\end{tabular}
}
\end{table}

\textbf{Cross-platform performance correlation.}
Several perceptual dimensions show cross-platform ranking agreement, while functional performance exhibits a less consistent simulator-to-real relationship (Figures~\ref{fig:corr-task}--\ref{fig:corr-libe-real}). The observed RoboTwin-to-real task-success association is weak, and low physical success rates highlight the difficulty of transferring functional utility across platforms. These findings motivate direct physical evaluation alongside simulation; correlation estimates and small-sample interpretation are detailed in Appendix~\ref{sec:crossplatform_stats}.

\section{Limitations}\label{limitation}
Our benchmark is explicitly designed as a broad diagnostic stress test rather than a demonstration of turn-key robotic deployment. To prioritize benchmarking breadth across 12 distinct models, 3 modalities, and multiple platforms under heavy compute constraints (Appendix~\ref{sec:compute}), our primary RL environment suite operates under standardized fixed-seed protocols per configuration. Furthermore, physical evaluations ($n=10$ to $20$ trials per condition) serve as diagnostic probes to quantify transfer gaps under hardware noise. These experiments establish a rigorous baseline for identifying systemic failure modes, while broad, closed-loop physical deployment remains an open challenge requiring future advances in contact dynamics (Appendix~\ref{sec:limitations_details}).

\section{Conclusion and Future Work}

We introduce WorldArena 2.0, a benchmark built on WorldArena that extends embodied world model evaluation across modality, functionality, and platform. Visuotactile evaluation exposes task-dependent strengths and failures, learned-environment training improves simulator policy performance, and cross-platform comparisons reveal gaps in physical task competence. These complementary evaluations connect predictive quality with practical embodied utility. The current physical studies remain preliminary, motivating broader real-world and closed-loop evaluation in future versions.

\clearpage

\bibliographystyle{unsrt}
\bibliography{ref}

\clearpage
\appendix


\clearpage
\section{Experimental Platforms and Tasks}\label{platform}

\FloatBarrier
\subsection{RoboTwin 2.0}
\textbf{RoboTwin 2.0}~\citep{chen2025robotwin} is a scalable bimanual simulation environment comprising 731 objects across 147 categories, featuring extensive domain randomization (e.g., scene clutter, lighting, textures, table height, and language instructions). We evaluate two tasks that benchmark the world models' ability to simulate rich interaction dynamics: \textit{Adjust Bottle} and \textit{Click Bell}. \textit{Adjust Bottle} requires the robot to reorient and position a bottle, testing fine-grained pose adjustment and grasp stability prediction. \textit{Click Bell} involves pressing a bell to produce an audible click, requiring the world model to accurately capture contact-rich dynamics and precise force application. We disable domain randomization during RL checkpoint evaluation and evaluate policies within the selected task distribution. The systematic randomization also exposes visual and spatial distribution shifts, making RoboTwin 2.0 a stress test for robustness beyond memorized backgrounds or object poses.

\FloatBarrier
\subsection{LIBERO}
\textbf{LIBERO}~\citep{liu2023libero} provides 130 language-conditioned single-arm manipulation tasks built upon Robosuite~\citep{zhu2020robosuite}, utilizing procedural generation to isolate spatial, object, goal, and mixed knowledge shifts. We select \textit{Turn on the Stove} as our evaluation task. This task enables a fine-grained assessment of a world model's ability to capture both object-relational and articulated-body dynamics. We include LIBERO because it decouples different types of knowledge transfer, allowing us to diagnose precisely which aspects of environment dynamics a world model struggles to learn. This structured diagnostic capability is absent in purely randomization-based benchmarks and is critical for understanding a model's failure modes.

\FloatBarrier
\subsection{AgileX Split-Type ALOHA}
\textbf{AgileX Split-Type ALOHA}~\citep{agilex2025aloha} serves as the real-world robot testbed. This platform features a master-follower teleoperation architecture built on the RANGER MINI 3.0 chassis, equipped with PiPER 6-DoF lightweight robotic arms. For the cross-platform evaluation in Sec \ref{sec:cross-plat}, we use two real-world tasks, \textit{Pour Water}, which tests the world model's ability to predict fluid dynamics and deformable material behaviors under gravity; and \textit{Wipe Table}, which evaluates long-horizon contact sequences and friction modeling across extended surface trajectories, with ten trials per condition. Separate physical studies for the tactile and WM-RL extensions use their own tasks and protocols, described in Appendices \ref{sec:tactile_physical} and \ref{sec:additional_diagnostics}, respectively. Our physical evaluation uses small object-reset displacements. Physical trials retain natural lighting and background variation, providing a direct check of sim-to-real performance. The ALOHA platform brings the evaluation to physical reality, where the world model must contend with unmodeled effects such as sensor noise, variable friction, and imperfect actuation. 

\FloatBarrier
\subsection{Real-World Task Definitions}
\begin{figure}[H]
    \centering
    \includegraphics[width=0.55\textwidth]{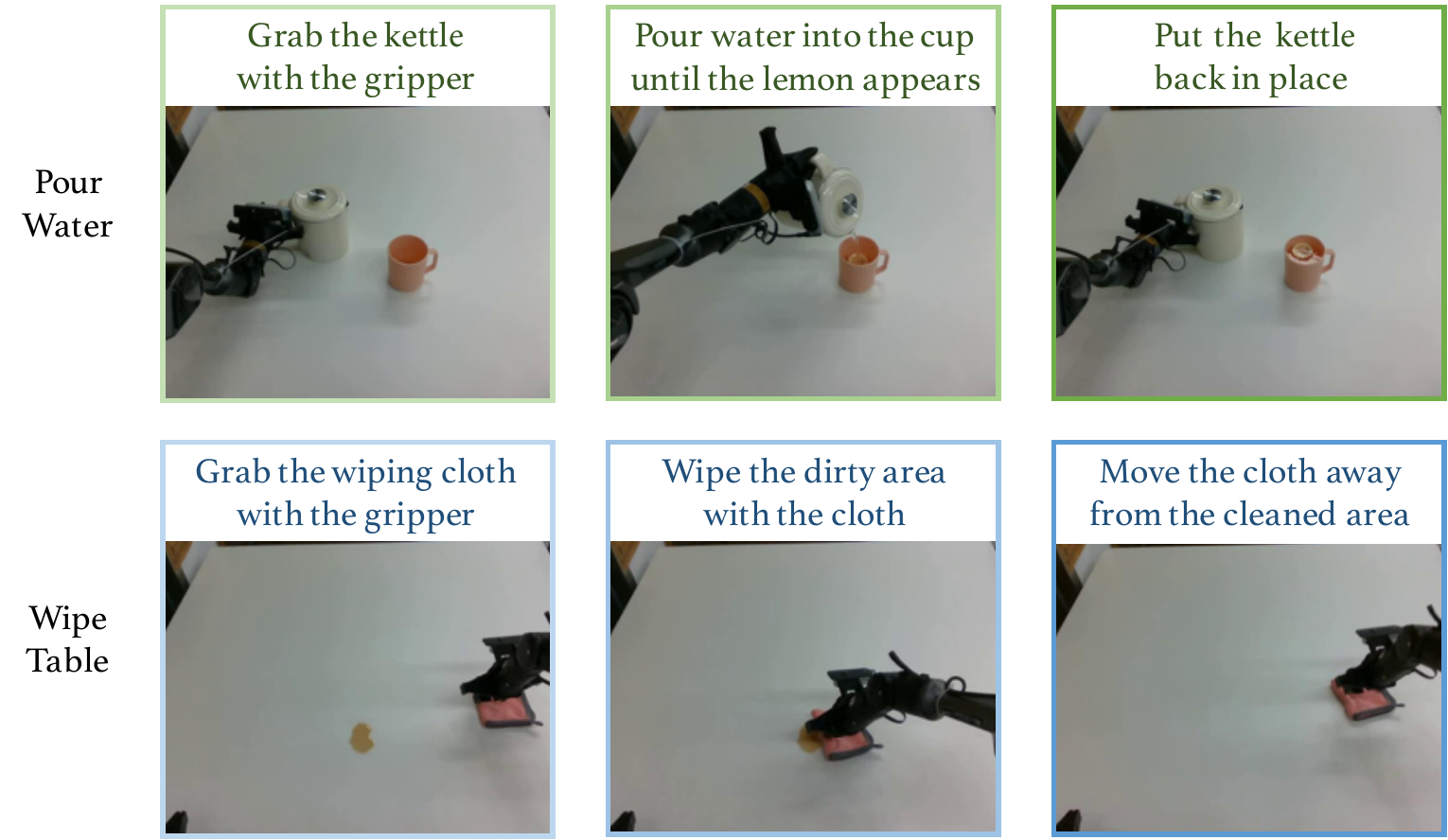}
    \caption{Illustration of the two real-world embodied tasks.}
    \label{fig:real_task}
\end{figure}

The Pour Water and Wipe Table tasks are illustrated in Figure~\ref{fig:real_task}.

\section{Supplementary Details for the Modality Extension}\label{sec:modality_appendix}

\FloatBarrier
\subsection{Visuotactile Evaluation Setup}
The evaluation uses the official UniVTAC dataset and the modular pipeline in Figure~\ref{fig:vtwm}. A tactile VAE encodes tactile deformation sequences into the video latent space; a two-stream world model jointly predicts visual and tactile observations; an action diffusion head converts the predicted latents into actions. The simulated tasks are \textit{Insert HDMI} and \textit{Lift Bottle}.

\FloatBarrier
\subsection{Conditioning and Decoder Interface}\label{sec:interfaces}
The world model receives visual and tactile context together with task text. The action decoder receives past states and actions and predicted visuotactile latents. Decoders are trained separately for each world model under common training settings. Past actions supplied to the decoder do not imply that the generator is conditioned on the executed future action chunk. Downstream success therefore measures the complete world-model--decoder system.

\FloatBarrier
\subsection{Simulated Visuotactile Evaluation Protocol}
Each visuotactile model is trained for 100 episodes. Tactile prediction quality is evaluated with PSNR and SSIM. Functional evaluation uses ten trials per model and task, with each trial capped at 200 environment steps. Table~\ref{tab:tactile_results} reports the point estimates; confidence intervals are provided below.

As an additional temporal diagnostic, contact-timing coverage is 97.2\% for Wan 2.2, 95.3\% for Vidar, and 87.6\% for Genie Envisioner. This alignment with task success demonstrates that contact-timing precision provides a critical dynamic diagnostic beyond static pixel fidelity, directly capturing the key phases of contact inception and release during manipulation.
\FloatBarrier
\subsection{Real-Robot Tactile Ablation}\label{sec:tactile_physical}
The comparison evaluates FastWAM~\citep{yuan2026fast} with and without the tactile branch while holding the remaining training and evaluation settings fixed. Each variant is trained for 200 episodes. Xense tactile sensing~\citep{xensewebsite} is used for two tasks, Chip and Wiping, with 20 evaluation trials per task, action chunk 25, a maximum of 600 steps, and no human intervention. We evaluate tactile input through task-level success rates in this FastWAM ablation.
\begin{table}[H]
\centering\small
\caption{Real-robot tactile evaluation. Success rates (\%) for FastWAM with and without tactile input.}
\label{tab:real_tactile_added}
\begin{tabular}{lccc}\toprule
Method & Chip & Wiping & Mean\\\midrule
FastWAM & 15 & 20 & 17.5\\
FastWAM+Tactile & 45 & 30 & 37.5\\\bottomrule
\end{tabular}
\end{table}

FastWAM achieves 15\% and 20\% success on Chip and Wiping, respectively; adding tactile input yields 45\% and 30\% (Table~\ref{tab:real_tactile_added}).

\FloatBarrier
\subsection{Statistical Uncertainty}\label{sec:trial_uncertainty}

For binary task outcomes with recoverable integer counts, we calculate two-sided 95\% Wilson score intervals~\citep{wilson1927probable}. With $\hat p=\frac{k}{n}$ and $z=1.96$, the endpoints are
\[
\frac{\hat p+\frac{z^2}{2n}\ \pm\ z\sqrt{\frac{\hat p(1-\hat p)}{n}+\frac{z^2}{4n^2}}}{1+\frac{z^2}{n}}.
\]
These intervals summarize trial-level uncertainty conditional on the evaluated policy and an independent-binomial approximation. They do not quantify seed variation, paired-method effects, or checkpoint-selection bias. The counts below are reconstructed from task success rates using 10 trials per UniVTAC condition and 20 per real tactile condition.
\begin{table}[htbp]\centering\small
\caption{95\% Wilson intervals (percent) for UniVTAC task success, with ten trials per task.}\label{tab:trial_intervals}
\begin{tabular}{lcc}\toprule
UniVTAC model ($n=10$/task) & Insert HDMI & Lift Bottle\\\midrule
ACT & 20 [5.7, 51.0] & 80 [49.0, 94.3]\\
Vidar & 70 [39.7, 89.2] & 0 [0.0, 27.8]\\
Genie Envisioner & 0 [0.0, 27.8] & 0 [0.0, 27.8]\\
Wan 2.2 & 100 [72.2, 100.0] & 0 [0.0, 27.8]\\\bottomrule
\end{tabular}\end{table}
\begin{table}[htbp]\centering\small
\caption{Real tactile evaluation counts and 95\% Wilson confidence intervals, corresponding to the point estimates in Table~\ref{tab:real_tactile_added}. The intervals describe each condition separately and do not constitute a test of the difference between conditions.}\label{tab:real_tactile_ci}
\begin{tabular}{llcc}\toprule
Method & Task & Successes/trials & Success (\%; 95\% CI)\\\midrule
FastWAM & Chip & 3/20 & 15 [5.2, 36.0]\\
FastWAM & Wiping & 4/20 & 20 [8.1, 41.6]\\
FastWAM+Tactile & Chip & 9/20 & 45 [25.8, 65.8]\\
FastWAM+Tactile & Wiping & 6/20 & 30 [14.5, 51.9]\\\bottomrule
\end{tabular}\end{table}

\section{Supplementary Details for the Functionality Extension}\label{sec:functionality_appendix}

\subsection{WM-RL Pipeline and Training Data}\label{sec:wm4rl_app}
To evaluate world models as RL training environments for embodied policies, we construct a unified end-to-end pipeline based on RLinf~\citep{zang2025rlinf,yu2025rlinf}. We select task \textit{Adjust Bottle} and \textit{Click Bell} from RoboTwin 2.0~\citep{chen2025robotwin}, with $\pi_{0.5}$~\citep{intelligence2025pi_} as the base embodied policy. We collect 3,000 simulator trajectories containing state, action, and visual-observation sequences using two policies with different performance levels and the simulator's native action planner; the same dataset is used to adapt the world and reward models. We then initialize policy optimization by fine-tuning $\pi_{0.5}$ with SFT on 1,000 high-quality expert trajectories generated by the planner. During RL optimization, the initialized policy interacts with each learned environment, and GRPO~\citep{shao2024deepseekmath} updates the policy under a fixed iteration budget. Finally, selected checkpoints are executed in the ground-truth RoboTwin 2.0 simulator, where the simulator task-success predicate determines task success.

\FloatBarrier
\subsection{Formalism and Component Interfaces}\label{sec:formalism}
We model the underlying task as a finite-horizon POMDP $\mathcal{M}=(\mathcal{S},\mathcal{O},\mathcal{A},P,\Omega,R,\rho_0)$, where $s_t\in\mathcal{S}$ is hidden state, $o_t\sim\Omega(\cdot\mid s_t)$ is an observation, $s_{t+1}\sim P(\cdot\mid s_t,a_t)$ is a transition, $r_t\sim R(\cdot\mid s_t,a_t)$ is the environment reward, and $\rho_0$ is the initial-state distribution. In the RL experiments, $P$ and $R$ denote dynamics and rewards. The video world model does not expose $s_t$; $\hat P_\phi$ is therefore a predictive kernel over future observations conditioned on the component-specific interaction history, rather than a parameterization of the latent-state kernel $P$. With instruction $g$ and history $h_t=(g,o_0,a_0,\ldots,o_t)$, component context maps $c_\pi,c_\phi,c_\psi$ make the shorthand in Eq.~\ref{eq:wm_recursion} explicit:
\begin{align}
a_t&\sim\pi_\theta(\cdot\mid c_\pi(h_t)),\\
\hat o_{t+1}&\sim\hat P_\phi(\cdot\mid c_\phi(h_t),a_t),\\
\hat r_t&=\hat R_\psi(c_\psi(h_t,a_t,\hat o_{t+1})).
\end{align}
The history is updated by appending $a_t$ and $\hat o_{t+1}$. Each context map selects the inputs its component actually supports; no full-history conditioning or Markov observation assumption is implied. The policy receives the head-camera observation, proprioceptive state, and task instruction. One model interaction uses an action chunk and predicts its corresponding future observations. Proxy and similarity rewards score a current image, whereas the VLM reward scores visual sequence context. For each sampled group of $G=4$ rollouts with reward scores $q_i$, GRPO~\citep{shao2024deepseekmath} forms group-relative advantages $\hat A_i=\frac{q_i-\bar q}{\operatorname{std}(q)+\epsilon}$ and uses them in the clipped policy objective, where $\epsilon$ is a numerical stabilizer. Optimization uses an actor-only loss without a critic model.

\FloatBarrier
\subsection{GRPO Optimization and Checkpoint Protocol}\label{sec:statistical_protocol}
The policy is $\pi_{0.5}$ (3.3B), initialized by SFT on 1,000 planner-generated expert trajectories. Only the action expert (approximately 0.3B parameters) is updated; the VLM backbone remains frozen. To preserve the pre-trained temporal inductive bias of each foundation model, we adhere to each architecture's native action-chunk and temporal interface rather than enforcing an artificial uniform chunking that could degrade latent dynamics. Consequently, we standardize on a uniform optimization iteration budget to reflect practical model-based adaptation regimes (Appendix~\ref{sec:compute}).

\begin{table}[H]
\centering
\small
\caption{Shared actor-only GRPO settings for WM-based policy optimization. Rollout settings vary by world model.}
\label{tab:rl_hyperparams}

\begin{tabularx}{\textwidth}{
@{}
>{\raggedright\arraybackslash}p{0.28\textwidth}
>{\raggedright\arraybackslash}X
@{}
}
\toprule
\textbf{Item} & \textbf{Setting}\\
\midrule

Optimization &
GRPO, group size 4; token-mean loss aggregation\\

Clipping &
Policy-ratio clip low/high $0.20/0.28$; dual clip $c=3.0$\\

Regularization &
KL coefficient 0; entropy bonus 0\\

Advantage &
Group-relative normalization within groups of four\\

Optimizer &
AdamW~\citep{loshchilov2019adamw}; weight decay 0.01; gradient clipping at 1.0\\

Action head &
Flow matching~\citep{lipman2022flow}; 5 denoising steps; flow-SDE noise 0.3;
14-D absolute joint actions\\

Policy input &
Head-camera observations, task instruction, and proprioceptive state\\

Checkpoint evaluation &
Fixed-seed RoboTwin episodes; no domain randomization\\

Selection &
Best success observed on these evaluation episodes\\

Training protocol \& seed & Fixed iteration budget; standardized single initialization seed (App.~\ref{sec:compute})\\

\bottomrule
\end{tabularx}
\end{table}

Following standard model-based RL evaluation conventions, checkpoints are periodically validated on fixed, held-out RoboTwin environments disjoint from the training trajectories. We report the peak policy performance achieved within the fixed training budget, establishing a standardized measure of the optimization upper bound enabled by each learned simulator.

\FloatBarrier
\subsection{Reward Designs}\label{sec:reward_designs}
\subsubsection{Proxy Reward}
The proxy uses a ResNet-18 visual backbone~\citep{he2016resnet}, frozen T5-base cross-attention~\citep{raffel2020t5}, and an MLP head, trained on the same 3,000-trajectory simulator dataset used to adapt the world models.
\subsubsection{VLM Reward}
The VLM is Qwen3.5-9B~\citep{team2026qwen3}, applied zero-shot to visual sequence context and served through SGLang~\citep{zheng2024sglang}.
\subsubsection{Similarity Reward}
The similarity design uses LPIPS similarity~\citep{zhang2018lpips} between a predicted observation and a task-goal observation.
\subsubsection{Simulator Baseline Reward}
Simulator-based RL uses the native RoboTwin task reward. The proxy, VLM, and similarity comparisons apply to policy optimization in learned world-model environments; SFT uses no RL reward. All policies are evaluated with the simulator's task-success criterion.

The simulator's success predicate uses internal spatial states that are unavailable for a generated video. Consequently, simulator rewards cannot simply be attached to arbitrary model-generated transitions without a validated state-reconstruction mechanism. The simulator-trained baseline changes the training dynamics, so it is not an oracle-reward-only ablation with the generator held fixed. Ground-truth simulator success is instead the evaluation outcome for every trained policy.

\FloatBarrier
\subsection{Reward Sensitivity Analysis}\label{sec:reward_correlation}
The three learned reward designs differ in supervision and computational cost. Within each reward comparison, the same reward model is applied to all generators; this does not imply equal calibration error across their image distributions.

\begin{table}[htbp]\centering\small
\caption{Correlations between success rates under different reward designs. Pearson $r$ measures linear agreement; Spearman $\rho$ measures rank agreement. The full set contains nine methods (including SFT and simulator RL) on two tasks. For this pooled analysis, the common SFT and simulator-RL baseline scores are replicated across reward-design comparisons because these baselines do not depend on the learned reward family. These are descriptive summaries of success rates after training, not direct reward-calibration measurements.}\label{tab:reward_correlations}
\begin{tabular}{llcc}\toprule
Subset & Reward pair & $r$ & $\rho$\\\midrule
All 18 points & Proxy--VLM & 0.916 & 0.846\\
 & Proxy--similarity & 0.971 & 0.977\\
 & VLM--similarity & 0.952 & 0.845\\\midrule
Click Bell (9 methods) & Proxy--VLM & 0.917 & 0.983\\
 & Proxy--similarity & 0.988 & 0.983\\
 & VLM--similarity & 0.952 & 0.967\\\midrule
Adjust Bottle (9 methods) & Proxy--VLM & 0.940 & 0.850\\
 & Proxy--similarity & 0.915 & 0.900\\
 & VLM--similarity & 0.972 & 0.833\\\bottomrule
\end{tabular}\end{table}

Pooling tasks can mix between-task and within-task differences, and including common SFT and high-performing simulator baselines can strengthen correlation. Task-specific coefficients therefore accompany the pooled analysis. Correlation does not demonstrate identical rankings, eliminate shared reward bias, or estimate uncertainty across independently trained policies. The similarity results are retained as sensitivity diagnostics only. The seven-world-model analysis in Table~\ref{tab:wm_only_correlations} provides the primary reward-sensitivity comparison because it excludes the replicated common baselines.

\begin{table}[htbp]\centering\small
\caption{Additional sensitivity analysis restricted to the seven world models (14 method--task points), excluding SFT and simulator RL. Computed from the displayed point estimates; it remains a pooled two-task descriptive analysis.}\label{tab:wm_only_correlations}
\begin{tabular}{lcc}\toprule Reward pair & Pearson $r$ & Spearman $\rho$\\\midrule
Proxy--VLM & 0.768 & 0.706\\
Proxy--similarity & 0.925 & 0.957\\
VLM--similarity & 0.824 & 0.710\\
\bottomrule\end{tabular}\end{table}

Even when isolating the seven world models and excluding the high-performing simulator baseline, the rankings maintain substantial rank concordance across different reward formulations (Table~\ref{tab:wm_only_correlations}, Spearman $\rho$ up to 0.957). This consistency verifies that the relative capability profile of each world model is robust to the choice of training supervision.

The correlations in Tables~\ref{tab:reward_correlations} and~\ref{tab:wm_only_correlations} are computed from the success rates in Table~\ref{tab:wm_rl}.

\FloatBarrier
\subsection{Rollout Horizon and Coverage}\label{sec:transfer_diagnostics}
The relevant recursive horizon is the number of world-model generations, not just the number of low-level environment steps. For a configured step limit $H$ and action chunk size $K$, the maximum number of model queries is $\lceil H/K\rceil$, with fewer queries when an episode terminates early. We match each policy action chunk to the corresponding world model's temporal interface; chunk sizes, step limits, and batch sizes therefore vary across models. Action chunking bounds recursive generation depth, where prediction errors can accumulate, but a configured query budget is not an empirical stability horizon. Reliability also depends on reward calibration and coverage of the states and actions visited during policy optimization.

We evaluate trained policies under ground-truth simulator dynamics using the simulator's task-success criterion. This separates evaluation from learned training rewards, although reward exploitation during training remains possible. Rollout stability, reward calibration, and on-policy coverage require further quantitative characterization.

\FloatBarrier
\subsection{Computational Cost}\label{sec:compute}
RL training uses two A800-80GB GPUs, one for the policy and one for the learned environment. Average per-GPU memory use is approximately 57 GB. Under the proxy reward, recorded per-iteration wall times range from 7.8 to 28.8 minutes across the profiled world models. These measurements characterize batched offline policy optimization rather than real-time robot control.

\FloatBarrier
\subsection{Policy Optimization Dynamics}\label{sec:optimization_dynamics}
\begin{figure}[H]
    \centering
    \includegraphics[width=0.85\textwidth]{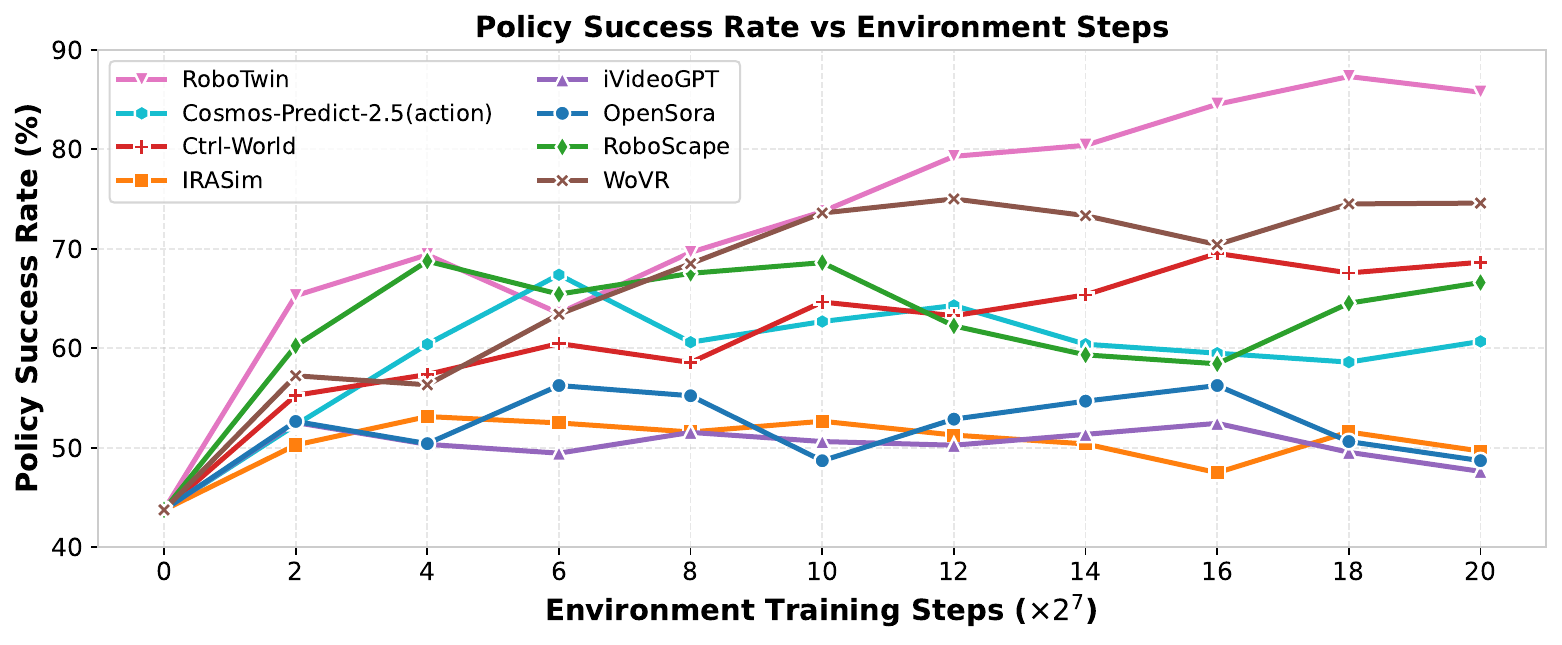}
    \caption{Checkpoint success rates on \textit{Click Bell} under the proxy reward over training. Each curve shows one training seed.}
    \label{fig:wm4rl-successrate}
\end{figure}

Figure~\ref{fig:wm4rl-successrate} illustrates the policy optimization trajectories on Click Bell across representative world-model environments. Across training steps, policies consistently exhibit stable reward progression, verifying that the learned dynamics reliably provide gradient signals for iterative policy improvement.

\FloatBarrier
\subsection{Future-Information Ablation and Attribution}\label{sec:future_ablation}
To test whether downstream action prediction genuinely uses the world-model future rather than relying only on history inputs, we perform an ablation on \textit{Adjust Bottle} using the per-WM action decoder. Providing ground-truth future observations yields 66\% success, compared with 32\% without future inputs and 33\% with shuffled futures. This confirms that temporally meaningful future information contributes substantially to downstream action prediction. This substantial performance gap (66\% with ground-truth futures vs. 33\% with shuffled futures) confirms the strong architectural sensitivity of the action decoder to temporal dynamics, ensuring that downstream execution effectively leverages the physical fidelity of predicted rollouts.

\FloatBarrier
\subsection{Real-Robot WM-RL Pilot}\label{sec:additional_diagnostics}
On AgileX ALOHA Lift Bottle, ten trials per policy produce progress summaries of 10\% (SFT), 25\% (WoVR-trained), and 30\% (Ctrl-World-trained). To capture fine-grained physical execution beyond coarse binary success, the real-world transfer study evaluates multi-stage task progress on Lift Bottle (tracking reaching, grasping, and lifting phases). Policies trained in world models achieve notable progress gains over SFT (25\% for WoVR and 30\% for Ctrl-World vs. 10\% for SFT), demonstrating meaningful transfer of learned interaction primitives to real hardware.

\section{Supplementary Details for the Platform Extension}\label{sec:platform_appendix}

\FloatBarrier
\subsection{Data Engine and Action Planner Protocols}\label{sec:platform_interfaces}
Table~\ref{tab:engine_planner} evaluates world models as data engines and action planners across RoboTwin 2.0, LIBERO, and AgileX ALOHA. The functional configurations use visual context and task text for future generation; the action decoder uses the generated future and its history inputs. Decoder parameters are trained separately per world model with common training settings. Task success measures the complete generator--decoder system.

\FloatBarrier
\subsection{Physical Evaluation Protocol}\label{sec:physical_protocol}
The Data Engine and Action Planner study uses ten trials per task/cell, chunk 45, at most 500 steps, and small object-reset displacements. The protocol permits only emergency-stop intervention. Failures include inaccurate grasping and incomplete stain removal. To ensure cross-hardware extensibility, the benchmark framework is built with modular abstraction layers, providing native interface support for additional robotic arms such as the Franka Research 3 (FR3) alongside the primary AgileX setup.

\FloatBarrier
\subsection{Extended Real-Robot Tasks and Qualitative Observations}
\label{sec:community_real_extension}

\paragraph{Task coverage and capabilities.}
The extended suite includes six vision-only tasks: \textit{Clean Table}, \textit{Instruction-Following Cleanup}, \textit{Fold Cardboard Box}, \textit{Fold Shirt}, \textit{Pour Water}, and \textit{Wipe Table}. Three visuotactile tasks complement these activities: \textit{Insert Two-Prong Plug}, \textit{Peel Cucumber}, and \textit{Pick Potato Chip}. Together, they probe instruction-conditioned object selection, deformable-object manipulation, sequential assembly, surface interaction, fragile-object handling, sustained tool contact, and precise insertion. Figure~\ref{fig:real_task_extended} illustrates the procedures using chronological demonstration frames, rather than model-evaluation rollouts. The original two-task illustration in Figure~\ref{fig:real_task} remains associated with the controlled cross-platform study.

\paragraph{Execution interface.}
The interface separates a remotely hosted policy, a scheduling hub, and a robot-side worker. Policies receive camera observations, robot states, and task instructions, and return action chunks for execution. Visuotactile configurations additionally expose tactile images and force/torque observations. The interface supports AgileX dual-arm joint control and Franka single-arm end-effector control; this support does not imply that every system was evaluated on both embodiments. Episodes are subject to a step budget, and successful completion requires avoiding safety interventions.

\paragraph{Qualitative observations.}
Available evaluations indicate uneven physical competence across tasks, with precision-demanding cardboard-box folding remaining particularly difficult. Even systems that sometimes completed box folding performed less reliably on this task than on wiping or garment folding. Box folding requires ordered manipulation of multiple panels and accurate edge alignment, making it a useful probe of fine manipulation. However, aggregate outcomes alone cannot determine whether failures originate in perception, planning, or contact control. Because the evaluated systems differ in training and implementation, these observations characterize task difficulty rather than establish a controlled model ranking. Demonstration frames illustrate the required procedures and should not be interpreted as evidence of an evaluated model's success.
\FloatBarrier
\begin{figure}[p]
    \centering
    \includegraphics[page=1,width=\textwidth]{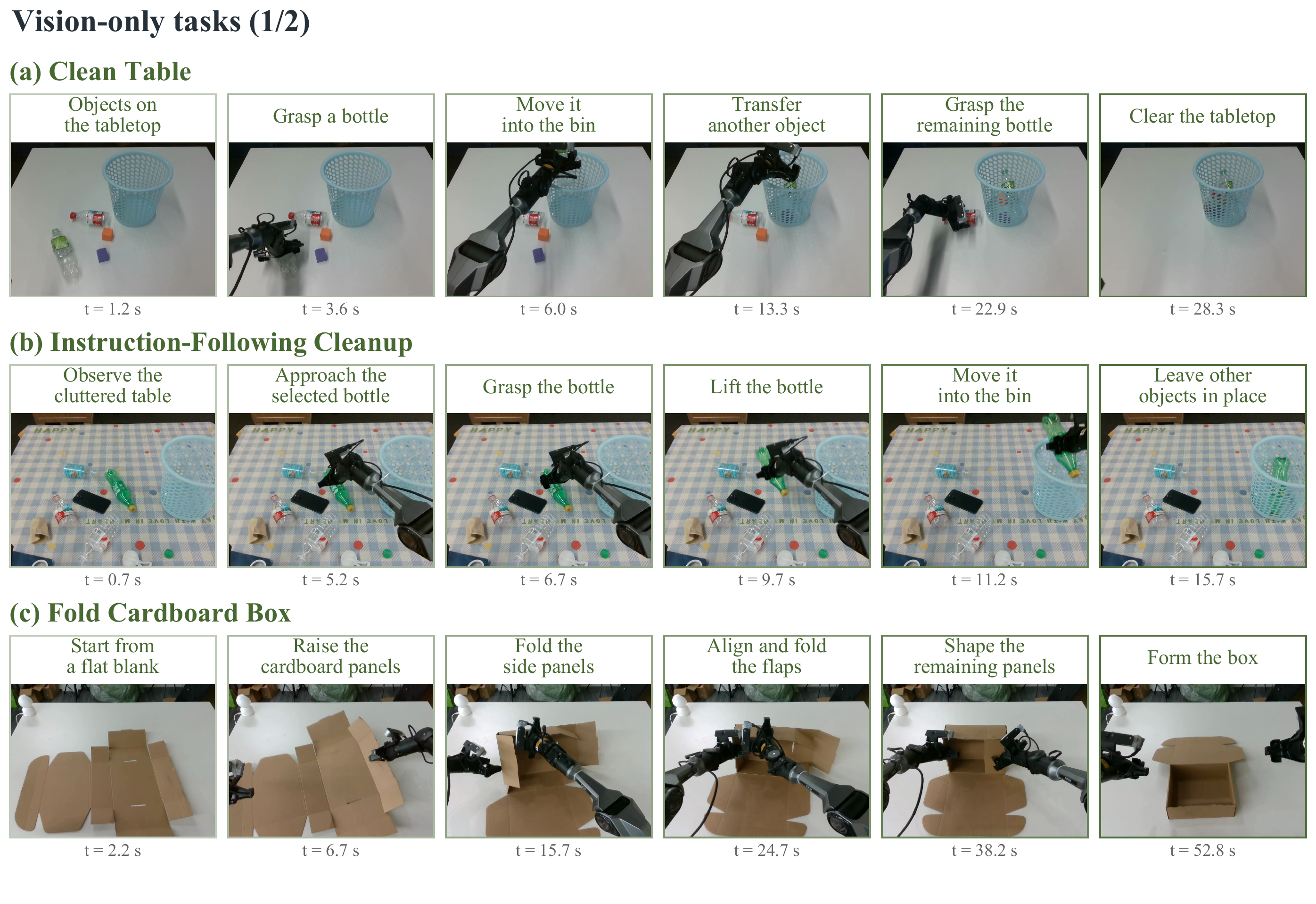}
    \caption{\textbf{Extended real-robot task suite (1/3): visual clearing and folding.} Tasks (a--c) show table clearing, instruction-following cleanup, and cardboard-box folding. Six chronological frames are shown per task. Timestamps indicate elapsed video time and are not aligned across tasks. Demonstration frames are selected and proportionally resized without generative editing; they illustrate procedures rather than model-evaluation outcomes. The remaining visual and visuotactile tasks appear on the following two pages.}
    \label{fig:real_task_extended}
\end{figure}

\begin{figure}[p]
    \ContinuedFloat
    \centering
    \includegraphics[page=2,width=\textwidth]{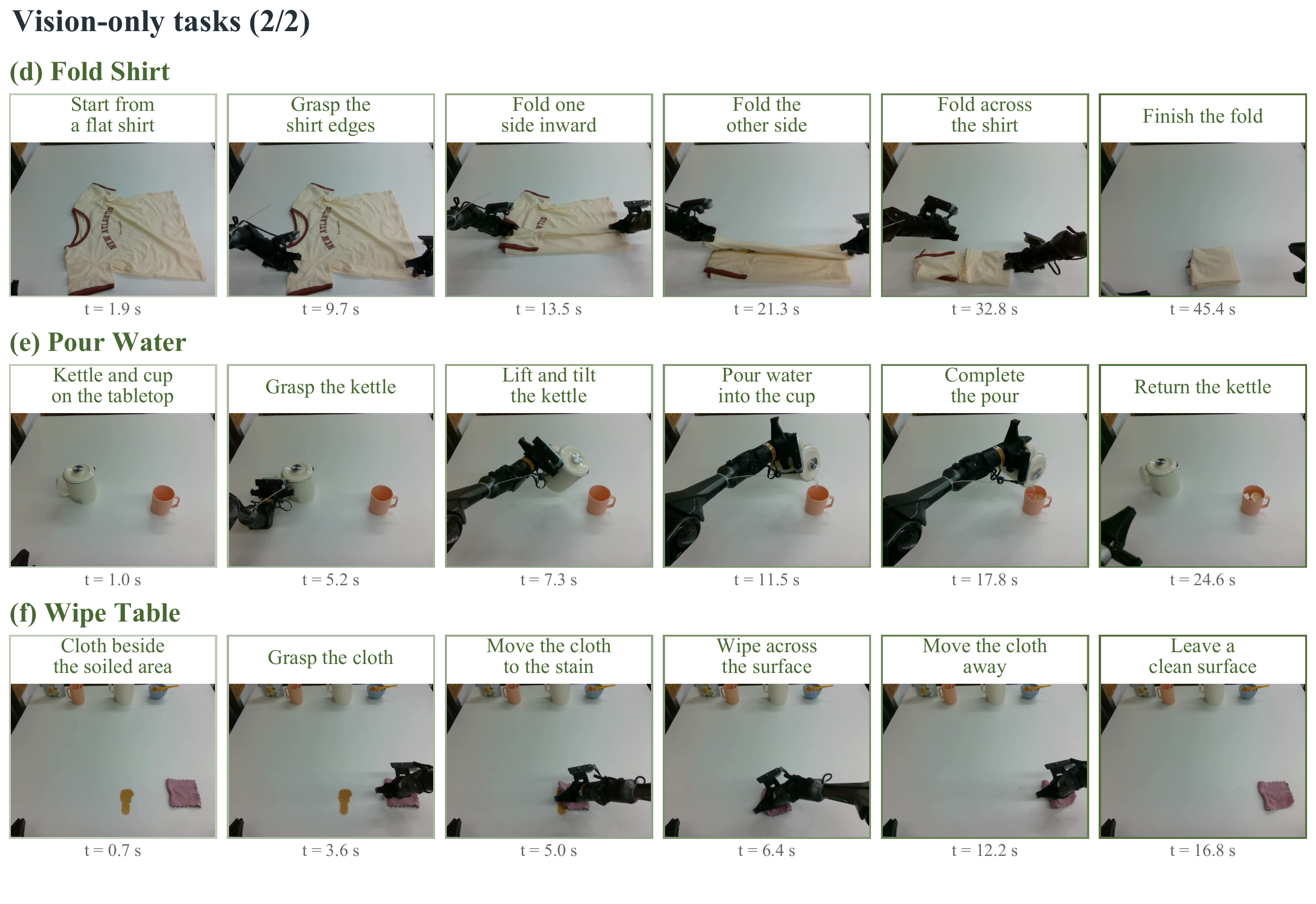}
    \caption{\textbf{Extended real-robot task suite (2/3): garment folding, pouring, and wiping.} Tasks (d--f) show shirt folding, water pouring, and table wiping, with six chronological frames per task, read left to right. All panels are vision-only task demonstrations, not outcomes attributed to a submitted model.}
    \label{fig:real_task_extended_visual_continued}
\end{figure}

\begin{figure}[p]
    \ContinuedFloat
    \centering
    \includegraphics[page=3,width=\textwidth]{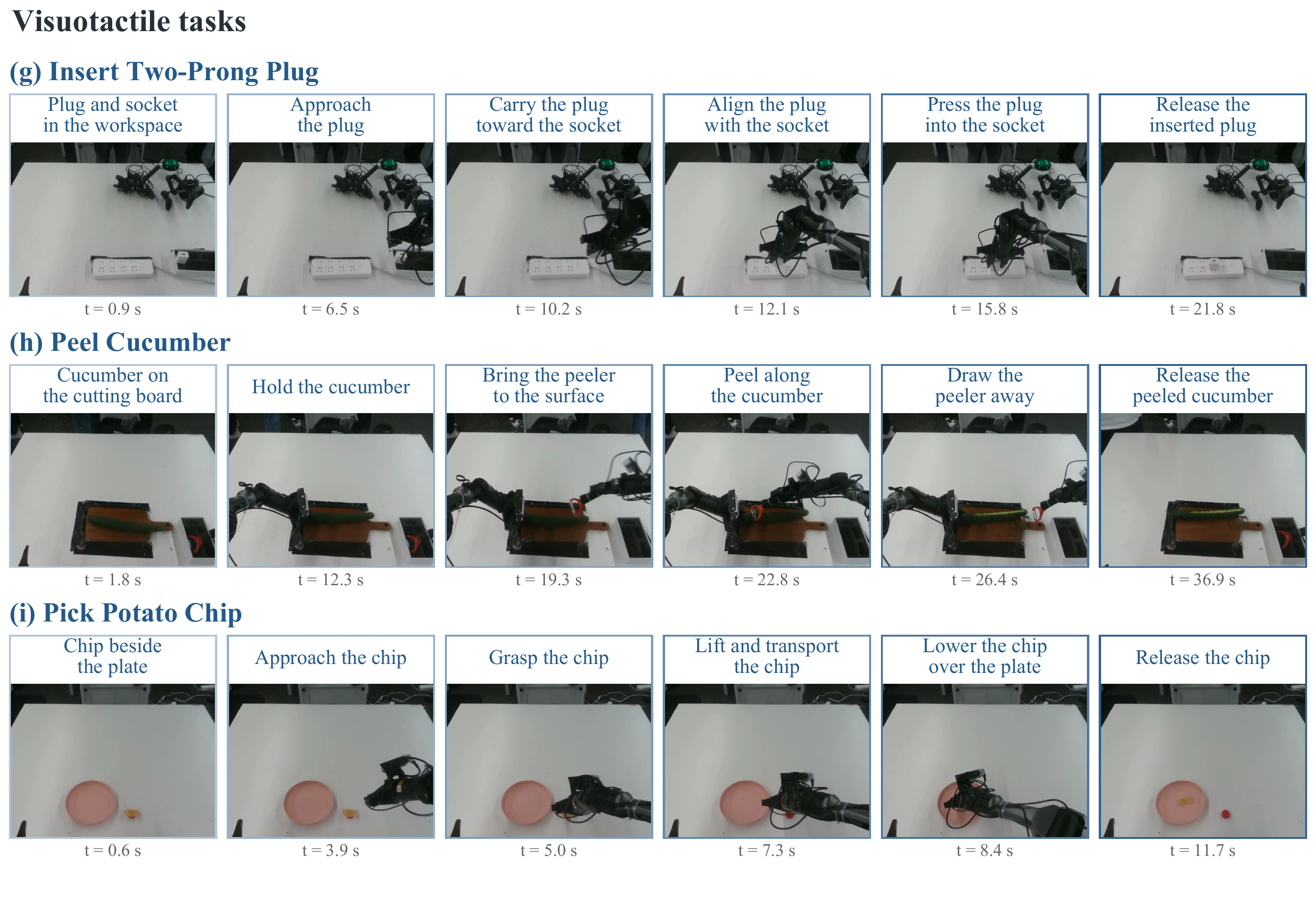}
    \caption{\textbf{Extended real-robot task suite (3/3): visuotactile manipulation.} Tasks (g--i) show plug insertion, cucumber peeling, and potato-chip transfer, with six chronological frames per task. The panels show the external RGB view of these visuotactile tasks, not tactile sensor images. These demonstrations illustrate task procedures and do not establish model success rates.}
    \label{fig:real_task_extended_tactile}
\end{figure}

\FloatBarrier

\FloatBarrier
\subsection{Physical Execution Cost}\label{sec:physical_cost}
In the action-planner setting, one generation-and-decoding cycle takes approximately 3.5--21 seconds per replan. The real-robot Data Engine and Action Planner experiments use action chunks of 45, approximately 1.5 seconds of inference, a 2-second control cycle, and 22 GB of GPU memory on A800 or 4090 hardware. These physical-pipeline measurements are not directly comparable with the batched RL timings.

\FloatBarrier
\subsection{Functional Evaluation and Uncertainty}\label{sec:platform_uncertainty}
The physical cells in Table~\ref{tab:engine_planner} use ten trials per condition. Table~\ref{tab:physical_intervals} maps the reported rates to 95\% Wilson intervals using the formula in Appendix~\ref{sec:trial_uncertainty}. These intervals describe individual conditions and do not account for training variability or paired-method effects. Small samples and zero-success outcomes limit ranking precision.

\begin{table}[htbp]\centering\small
\caption{95\% Wilson intervals for the real-world success rates in Table~\ref{tab:engine_planner}, with ten trials per condition.}\label{tab:physical_intervals}
\begin{tabular}{lcc}\toprule
Real-world reported rate & Count & 95\% interval\\\midrule
0\% & 0/10 & [0.0, 27.8]\\
10\% & 1/10 & [1.8, 40.4]\\
20\% & 2/10 & [5.7, 51.0]\\
30\% & 3/10 & [10.8, 60.3]\\
40\% & 4/10 & [16.8, 68.7]\\
50\% & 5/10 & [23.7, 76.3]\\\bottomrule
\end{tabular}\end{table}

\subsection{Cross-Platform Functional Correlation}\label{sec:crossplatform_stats}
\begin{figure*}[htbp]
    \centering
    
    \includegraphics[width=\textwidth]{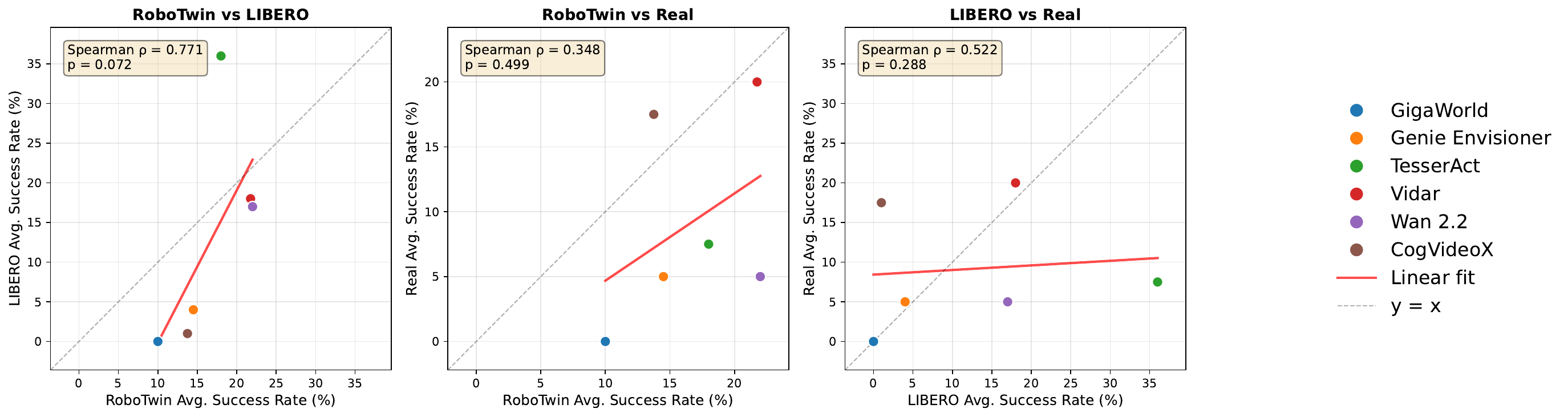}
    \caption{Cross-platform task success rate correlation between RoboTwin, LIBERO and the real-world robotic data.}
    \label{fig:corr-task}
\end{figure*}

RoboTwin--LIBERO (Figure~\ref{fig:corr-task}) shows a stronger positive association (\(\rho=0.771,p=0.072\)), whereas both simulator-to-real comparisons remain statistically uncertain (RoboTwin--Real: \(0.348,0.499\); LIBERO--Real: \(0.522,0.288\)). The small model set and many zero-success physical evaluations make this estimate uncertain. We treat the correlation as descriptive and assess real-world utility through direct physical evaluation.

\subsection{Full Perceptual Evaluation Results}\label{sec:perceptual_results}
Perceptual tables (Table~\ref{tab:vq_robotwin_part1}--~\ref{tab:vq_real_part2}) include both text- and action-conditioned configurations, explicitly including both Cosmos variants. Their output-only metrics describe the evaluated systems, but trajectory matching and controllability across heterogeneous conditioning are not controlled comparisons of transition fidelity. Accordingly, dynamics-quality comparisons should remain stratified by interface; functional success measures the complete generator--decoder system.

\FloatBarrier
\subsection{Cross-Platform Perceptual Correlations}\label{sec:perceptual_correlations}
\begin{figure*}[htbp]
    \centering
    \includegraphics[width=0.75\textwidth]{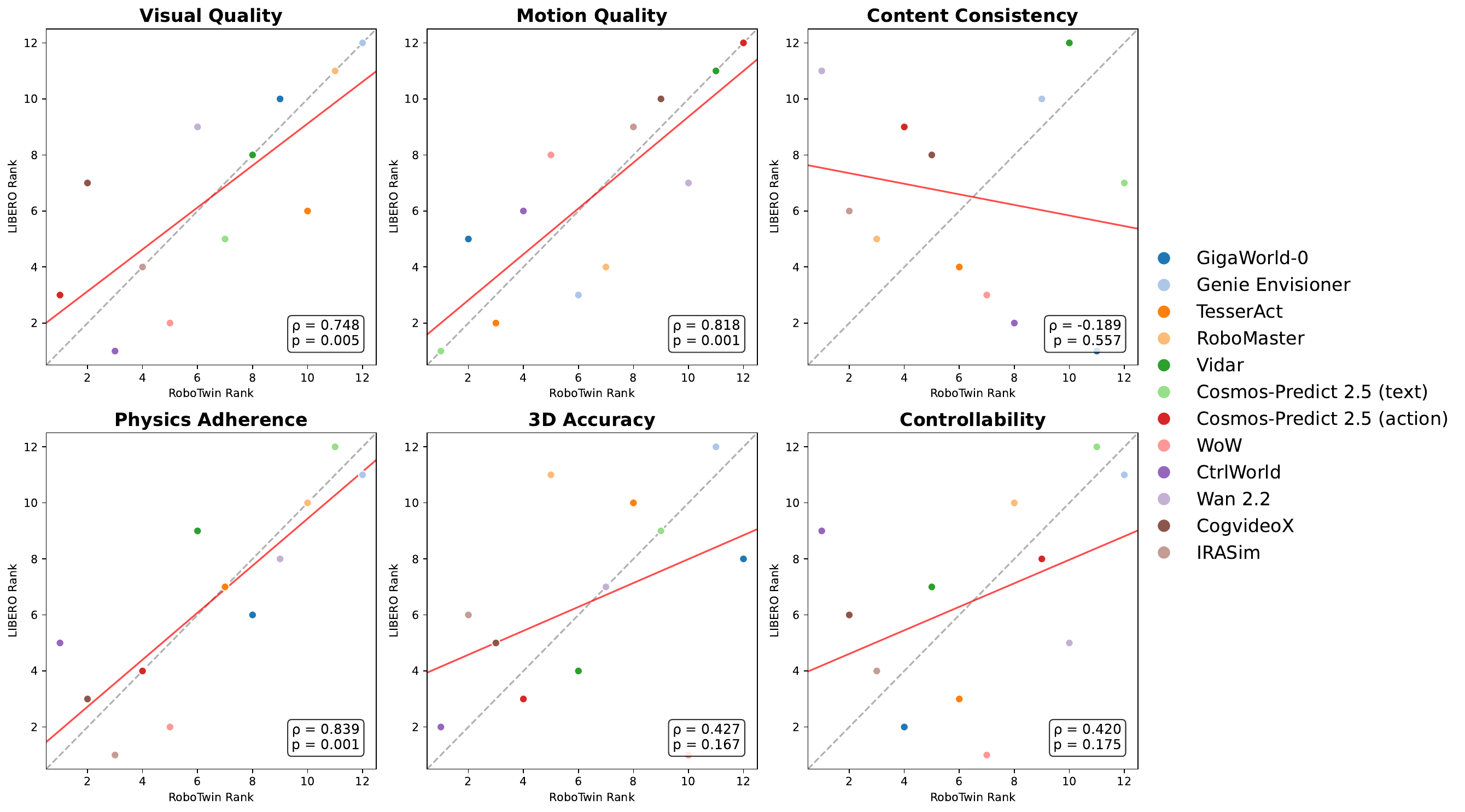}
    \caption{Cross-platform video quality correlation between RoboTwin and LIBERO.}
    \label{fig:corr-robo-libe}
\end{figure*}

\begin{figure*}[htbp]
    \centering
    \includegraphics[width=0.75\textwidth]{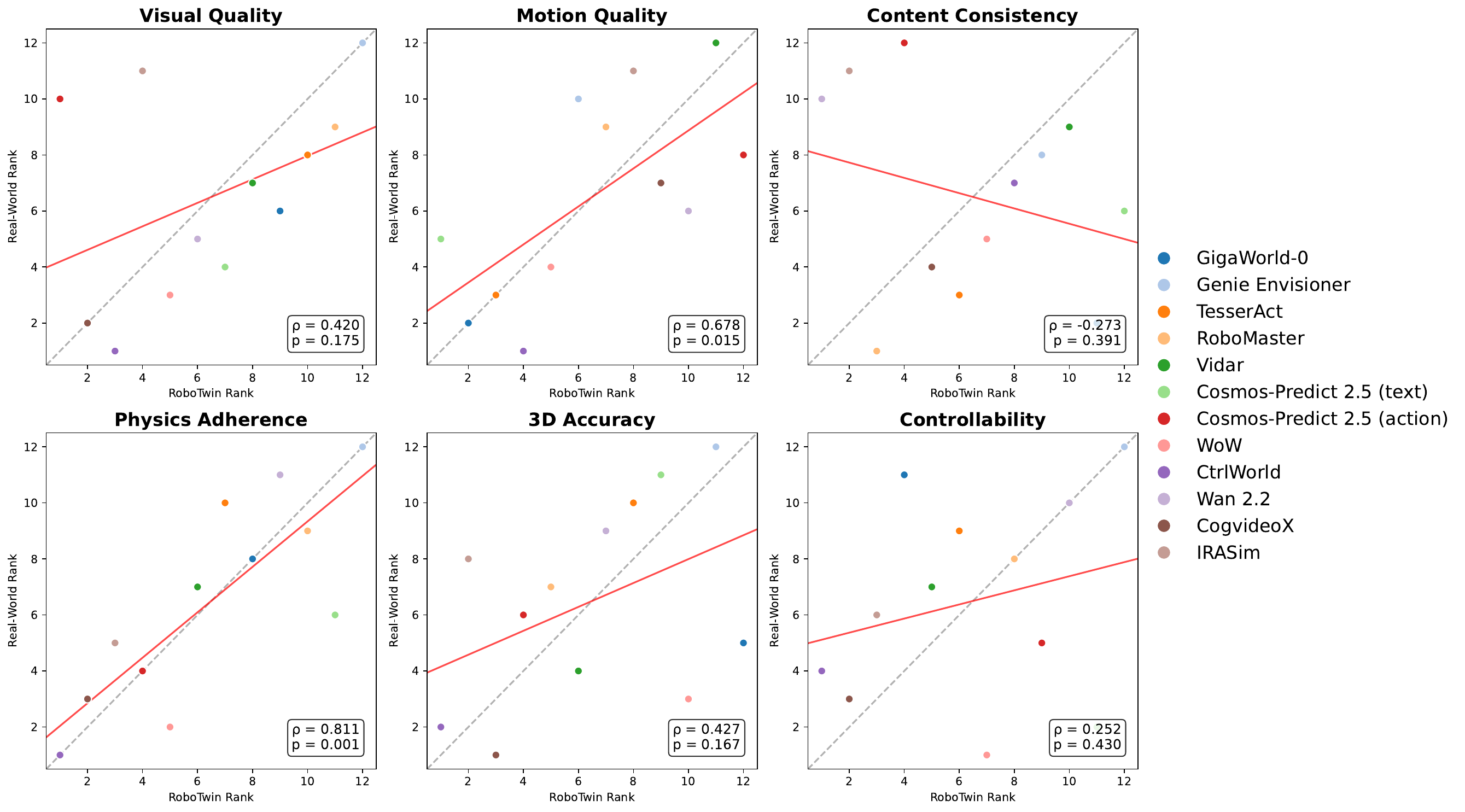}
    \caption{Cross-platform video quality correlation between RoboTwin and the real-world robotic data.}
    \label{fig:corr-robo-real}
\end{figure*}

\begin{figure*}[htbp]
    \centering
    \includegraphics[width=0.75\textwidth]{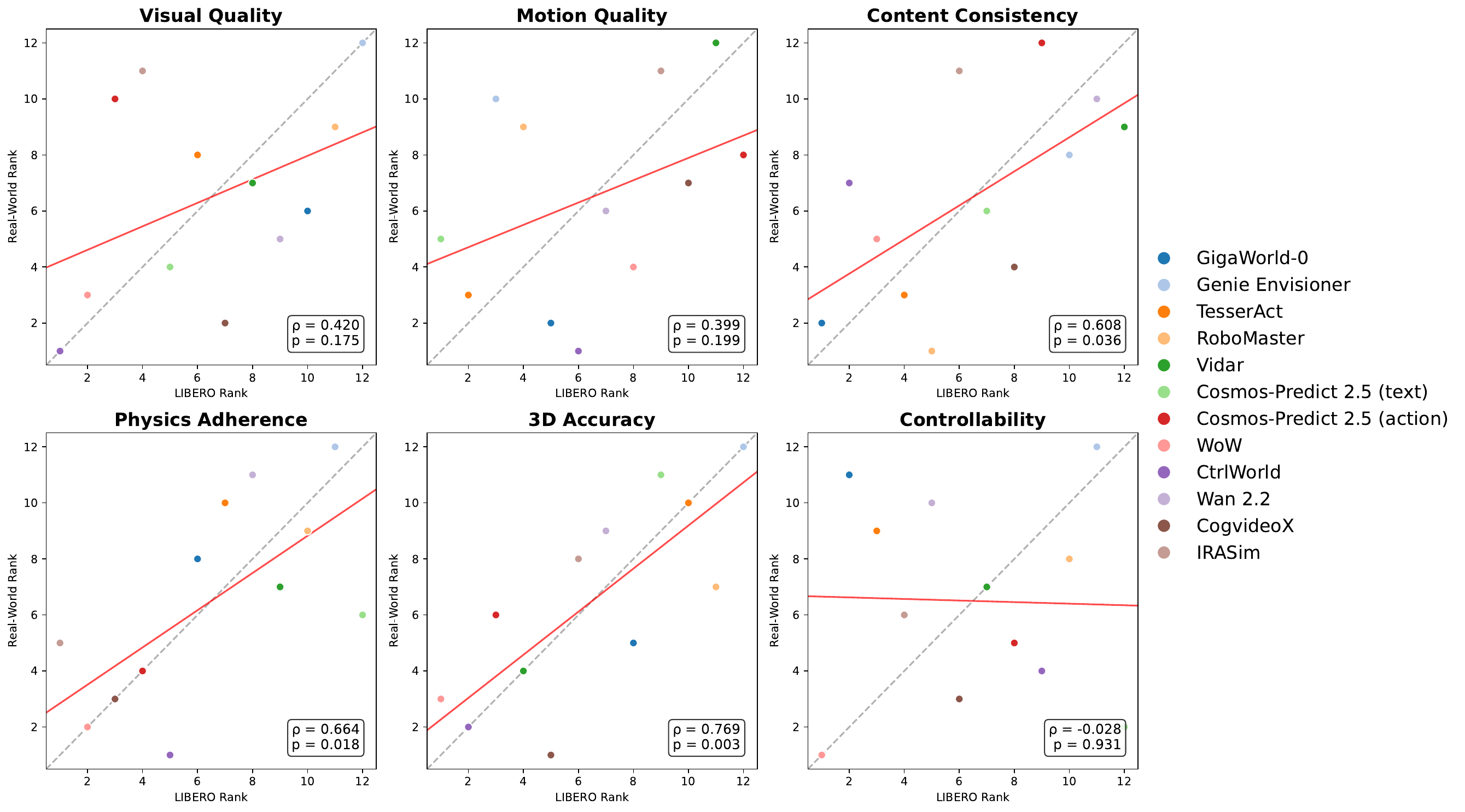}
    \caption{Cross-platform video quality correlation between  LIBERO and the real-world robotic data.}
    \label{fig:corr-libe-real}
\end{figure*}

Perceptual ranking agreement varies by dimension (Figures~\ref{fig:corr-robo-libe}--\ref{fig:corr-libe-real}). Motion quality, physics adherence, and 3D accuracy provide complementary diagnostics; their ability to predict physical task success requires further validation.

\FloatBarrier
\section{Extended Limitations and Responsible Use}\label{sec:limitations_details}

\subsection{Statistical Limitations}
Our RL benchmarking protocol balances extensive model coverage against foundation-scale compute demands (Appendix~\ref{sec:compute}), utilizing fixed evaluation environments to benchmark optimization ceilings. While scaling up seed replicates and physical trials will provide narrower estimation bounds, the current multi-platform protocol successfully exposes systematic performance discrepancies across embodiments and reward specifications.

\subsection{Modeling and Attribution}
For decoder-based functional protocols, success reflects the complete generator--decoder system; WM-RL results similarly reflect the coupled learned-environment--policy pipeline. Reward sensitivity and rollout reliability are discussed in Appendices~\ref{sec:reward_correlation} and~\ref{sec:transfer_diagnostics}. Generalization beyond the evaluated tasks, scenes, and embodiments remains to be tested.
\subsection{Computational Limitations}
Computational cost varies substantially across world models and reward implementations (Appendix~\ref{sec:compute}). The learned environments support offline policy optimization; derived replanning frequencies must be distinguished from physical low-level control rates. Open-loop action chunks constrain rapid contact feedback.

\subsection{Physical Safety and Supervision}
The non-tactile physical protocol permits only emergency-stop intervention, and the tactile pilot uses no human intervention. Physical deployment should retain human supervision and emergency-stop access and report all interventions. Because imperfect dynamics and rewards can induce unsafe actions, simulator gains and perceptual scores do not justify unsupervised deployment.

\section{Broader Impact}

WorldArena 2.0 evaluates embodied world models across modalities, functionalities, and platforms, exposing failures that can guide robust agents. Errors in learned dynamics can transfer to trained policies. Physical deployment therefore requires validation, intervention and safety-stop reporting, and human supervision.

\begin{table*}[h]
\centering
\caption{Video quality evaluation results across visual quality, motion quality and content consistency dimensions on Robotwin dataset.}
\label{tab:vq_robotwin_part1}
\small
\setlength{\tabcolsep}{4pt}
\resizebox{\textwidth}{!}{
\begin{tabular}{lccccccccc}
\toprule
\multirow{2}{*}{\textbf{\large Models}} & \multicolumn{3}{c}{\textbf{\large Visual Quality}} & \multicolumn{3}{c}{\textbf{\large Motion Quality}} & \multicolumn{3}{c}{\textbf{\large Content Consistency}} \\ 
\cmidrule(lr){2-4} \cmidrule(lr){5-7} \cmidrule(lr){8-10}
 & \begin{tabular}[c]{@{}c@{}}Image \\ Quality\end{tabular} & \begin{tabular}[c]{@{}c@{}}Aesthetic \\ Quality\end{tabular} & \begin{tabular}[c]{@{}c@{}}JEPA \\ Similarity\end{tabular} & \begin{tabular}[c]{@{}c@{}}Dynamic \\ Degree\end{tabular} & \begin{tabular}[c]{@{}c@{}}Flow \\ Score\end{tabular} & \begin{tabular}[c]{@{}c@{}}Motion \\ Smoothness\end{tabular} & \begin{tabular}[c]{@{}c@{}}Subject \\ Consistency\end{tabular} & \begin{tabular}[c]{@{}c@{}}Backgrd. \\ Consistency\end{tabular} & \begin{tabular}[c]{@{}c@{}}Photom. \\ Consistency\end{tabular} \\ 
\midrule
GigaWorld-0 &0.5041  &0.3991  &0.4413  &0.6709  &0.3118  &0.7811  &0.7303  &0.8563  &0.1756  \\
Genie Envisioner &0.2305  & 0.3289 &0.3340  &0.6930  &0.0855  &0.6966  &0.7760  &0.9024  &0.2006  \\
TesserAct&0.3322  &0.4590  &0.4579  &0.5150  &0.2447  &0.7579  &0.8250  &\textbf{0.9238}  &0.2491  \\
RoboMaster~\citep{robomaster2025} &0.3487  & 0.3842 & 0.2966 &0.6124  & 0.1484 &0.6940  & 0.8295 &0.9123  &0.3356  \\
Vidar &0.4145  &0.4068  &0.5608  & 0.2767 &0.1426  & 0.7973 & 0.7629 &0.8300  &0.2350  \\
Cosmos-Predict 2.5 (text)~\citep{nvidia2025cosmospredict25} &0.6668  &0.4501  &0.3126  &0.5911  &0.4302  &0.7882  &0.7488  &0.8511  &0.1383  \\
Cosmos-Predict 2.5 (action) &0.4489  &0.3576  & 0.9296 &0.3994  &0.0573  &0.7100  &0.8197  &0.8894  & 0.3528 \\
WoW &0.4587  &0.3868  &0.7440  &0.4608  &0.2706  &0.7692  &0.8161  &0.9025  &0.2170  \\
CtrlWorld &0.3522  &0.3893  &0.9185  &0.4257  &0.3449  &0.7377  &\textbf{0.8411}  &0.9057  &0.1729  \\
Wan 2.2 &0.3884  &0.3963  &0.7575  & 0.4349 &0.1269  &0.7019  &0.8388  &0.9042  &\textbf{0.4776}  \\
CogvideoX &0.3582  &0.3777  & \textbf{0.9384} &0.3166  &0.2189  &0.7391  &0.8083  & 0.8773 &0.3580  \\
IRASim &0.3489  &0.3623 &0.9330  &0.4139  &0.2083 &0.7052 & 0.8312 &0.9068 &0.3522\\
Veo 3.1 & 0.6605 &\textbf{0.4632}  &0.5694  &0.5450  & 0.1396 &0.6989  &0.7878  &0.8710  &0.3247  \\
Wan 2.6 &\textbf{0.6824}  &0.4433  &0.7229  &\textbf{0.7421}  &\textbf{0.4532}&\textbf{0.8539}  & 0.7517 &0.8687  &0.1904    \\
\bottomrule
\end{tabular}
}
\end{table*}

\begin{table*}[h]
\centering
\caption{Video quality evaluation results across physics adherence, 3D accuracy and controllability dimensions on Robotwin dataset.}
\label{tab:vq_robotwin_part2}
\small
\resizebox{\textwidth}{!}{
\begin{tabular}{lcccccccc}
\toprule
\multirow{2}{*}{\textbf{\large Models}} & \multicolumn{2}{c}{\textbf{\large Physics Adherence}} & \multicolumn{2}{c}{\textbf{\large 3D Accuracy}} & \multicolumn{3}{c}{\textbf{\large Controllability}} \\ 
\cmidrule(lr){2-3} \cmidrule(lr){4-5} \cmidrule(lr){6-8}
 & \begin{tabular}[c]{@{}c@{}}Interaction \\ Quality\end{tabular} & \begin{tabular}[c]{@{}c@{}}Trajectory \\ Accuracy\end{tabular} & \begin{tabular}[c]{@{}c@{}}Depth \\ Accuracy\end{tabular} & Perspectivity & \begin{tabular}[c]{@{}c@{}}Instruction \\ Following\end{tabular} & \begin{tabular}[c]{@{}c@{}}Semantic \\ Alignment\end{tabular} & \begin{tabular}[c]{@{}c@{}}Action Response \\Sensitivity\end{tabular} \\ 
\midrule
GigaWorld-0 &0.5368  &0.1552 &0.6316  &0.7596  &0.6156 &0.8591  &0.1134  \\
Genie Envisioner &0.2052 &0.0679 &0.8663  &0.5284  &0.2028 &0.8544  &0.0109  \\
TesserAct&0.5800  &0.1396 &0.7159  &0.7920  &0.6152 &0.8783  &0.0311  \\
RoboMaster &0.5364  &0.1158 &0.8335  &0.7588  &0.5772 &0.8761  &0.0352  \\
Vidar &0.5348  &0.1928 &0.7872  &0.7592  &0.5912 &0.8826  &0.0819  \\
Cosmos-Predict 2.5 (text) &0.3872  &0.0816 & 0.7051 &0.7964  &0.2664 & 0.7733 &\textbf{0.1418}  \\
Cosmos-Predict 2.5(action) &0.5500  &0.2945 &0.8862  &0.7644  &0.5840 &0.8879  &0.0133  \\
WoW &0.5564  &0.2058 &0.7283  &0.7672  &0.5692 &0.8842  & 0.0434 \\
CtrlWorld &0.6212  &\textbf{0.4766} &0.9300  &0.7960  &0.7272 & 0.8912 &0.0210  \\
Wan 2.2 &0.5184  &0.1627 &0.7768  &0.7660  &0.5376 &0.8877  & 0.0512 \\
CogvideoX &0.5940  &0.3526 & 0.9097 &0.7828  &0.7268 &\textbf{0.8977}  &0.0076  \\
IRASim & 0.5656 &0.3639 &\textbf{0.9312}  &0.7788 &0.6604 &0.8849&0.0526  \\
Veo 3.1 &\textbf{0.7872}  &0.1231 &0.7421  &\textbf{0.8276}  &\textbf{0.9328} & 0.8607 &0.0852  \\
Wan 2.6 & 0.7280 &0.1182 &0.7144  &0.8032  &0.8536 &0.8728  & 0.0992 \\
\bottomrule
\end{tabular}
}
\end{table*}

\begin{table*}[h]
\centering
\caption{Video quality evaluation results across visual quality, motion quality and content consistency dimensions on Libero dataset.}
\label{tab:vq_libero_part1}
\small
\setlength{\tabcolsep}{4pt}
\resizebox{\textwidth}{!}{
\begin{tabular}{lccccccccc}
\toprule
\multirow{2}{*}{\textbf{\large Models}} & \multicolumn{3}{c}{\textbf{\large Visual Quality}} & \multicolumn{3}{c}{\textbf{\large Motion Quality}} & \multicolumn{3}{c}{\textbf{\large Content Consistency}} \\ 
\cmidrule(lr){2-4} \cmidrule(lr){5-7} \cmidrule(lr){8-10}
 & \begin{tabular}[c]{@{}c@{}}Image \\ Quality\end{tabular} & \begin{tabular}[c]{@{}c@{}}Aesthetic \\ Quality\end{tabular} & \begin{tabular}[c]{@{}c@{}}JEPA \\ Similarity\end{tabular} & \begin{tabular}[c]{@{}c@{}}Dynamic \\ Degree\end{tabular} & \begin{tabular}[c]{@{}c@{}}Flow \\ Score\end{tabular} & \begin{tabular}[c]{@{}c@{}}Motion \\ Smoothness\end{tabular} & \begin{tabular}[c]{@{}c@{}}Subject \\ Consistency\end{tabular} & \begin{tabular}[c]{@{}c@{}}Backgrd. \\ Consistency\end{tabular} & \begin{tabular}[c]{@{}c@{}}Photom. \\ Consistency\end{tabular} \\ 
\midrule
GigaWorld-0 & 0.3056 & 0.4860 & 0.1693 & 0.2387 & 0.0814 & 0.5886 & \textbf{0.9082} & \textbf{0.9481} & 0.7859 \\
Genie Envisioner & 0.3688 & 0.2518 & 0.0303 & 0.3330 & 0.2273 & 0.5776 & 0.6867 & 0.8670 & 0.2074 \\
TesserAct & 0.3779 & \textbf{0.5249} & 0.3524 & \textbf{0.5060} & 0.1433 & 0.6917 & 0.8659 & 0.9023 & 0.3294 \\
RoboMaster & 0.3530 & 0.3883 & 0.0468 & 0.3729 & 0.1263 & 0.5780 & 0.9080 & 0.9347 & 0.1964 \\
Vidar & 0.3071 & 0.4881 & 0.3725 & 0.0986 & 0.0342 & 0.5026 & 0.3607 & 0.3705 & 0.4354 \\
Cosmos-Predict 2.5 (text) & \textbf{0.7406} & 0.5014 & 0.0205 & 0.3174 & \textbf{0.6573} & \textbf{0.9590} & 0.6578 & 0.8210 & 0.3818 \\
Cosmos-Predict 2.5 (action) &0.3642  &0.5064  &0.4205  &0.1150  &0.0374  &0.4465  &0.5550  &0.5742  &0.6496  \\
WoW & 0.3573 & 0.5061 & 0.4612 & 0.1972 & 0.0488 & 0.4773 & 0.7843 & 0.8184 & \textbf{0.8629} \\
CtrlWorld & 0.3530 & 0.4996 & \textbf{0.6087} & 0.2358 & 0.0698 & 0.5467 & 0.8396 & 0.8697 & 0.8227 \\
Wan 2.2 & 0.3197 & 0.4753 & 0.2919 & 0.1341 & 0.0525 & 0.5597 & 0.3954 & 0.4253 & 0.4681 \\
CogvideoX &0.3044  &0.4614  &0.4513  &0.1334  &0.0429  &0.4793  &0.5852  &0.6036  &0.6631  \\
IRASim &0.3484  &0.4843  &0.4407  &0.1364  &0.0442  &0.5012  &0.5633  &0.5941  &0.7352  \\
\bottomrule
\end{tabular}
}
\end{table*}

\begin{table*}[h]
\centering
\caption{Video quality evaluation results across physics adherence, 3D accuracy and controllability dimensions on Libero dataset.}
\label{tab:vq_libero_part2}
\small
\resizebox{\textwidth}{!}{
\begin{tabular}{lcccccccc}
\toprule
\multirow{2}{*}{\textbf{\large Models}} & \multicolumn{2}{c}{\textbf{\large Physics Adherence}} & \multicolumn{2}{c}{\textbf{\large 3D Accuracy}} & \multicolumn{3}{c}{\textbf{\large Controllability}} \\ 
\cmidrule(lr){2-3} \cmidrule(lr){4-5} \cmidrule(lr){6-8}
 & \begin{tabular}[c]{@{}c@{}}Interaction \\ Quality\end{tabular} & \begin{tabular}[c]{@{}c@{}}Trajectory \\ Accuracy\end{tabular} & \begin{tabular}[c]{@{}c@{}}Depth \\ Accuracy\end{tabular} & Perspectivity & \begin{tabular}[c]{@{}c@{}}Instruction \\ Following\end{tabular} & \begin{tabular}[c]{@{}c@{}}Semantic \\ Alignment\end{tabular} & \begin{tabular}[c]{@{}c@{}}Action Response \\Sensitivity\end{tabular} \\ 
\midrule
GigaWorld-0 & 0.6225 & 0.3690 & 0.8272 & 0.8090 & 0.7281 & 0.9102 & 0.0100 \\
Genie Envisioner & 0.2000 & 0.1495 & 0.7520 & 0.4920 & 0.2000 & 0.8758 & 0.0187 \\
TesserAct & 0.6000 & 0.3381 & 0.7039 & 0.7960 & 0.6380 & 0.9029 & 0.0185 \\
RoboMaster & 0.4060 & 0.1984 & 0.7918 & 0.6500 & 0.3580 & 0.9021 & 0.0363 \\
Vidar & 0.4100 & 0.2810 & 0.9133 & 0.7940 & 0.3820 & 0.9179 & 0.0389 \\
Cosmos-Predict 2.5 (text) & 0.2340 & 0.0015 & 0.5853 & \textbf{0.9400} & 0.2000 & 0.7637 & \textbf{0.1226} \\
Cosmos-Predict 2.5 (action) &0.5600  &0.4552  &0.9236  &0.8020  &0.4200  &0.9131  &0.0017  \\
WoW  & \textbf{0.6500} & 0.5859 & 0.9381 & 0.8060 & \textbf{0.7900} & 0.9147&0.0121 \\
CtrlWorld & 0.5160 & 0.4824 & \textbf{0.9431} & 0.7920 & 0.4040 & 0.9212 &0.0087  \\
Wan 2.2 & 0.5360 & 0.2467 & 0.8901 & 0.7960 & 0.5180 & 0.9102 & 0.0364 \\
CogvideoX &0.5200  &0.6676  &0.9267  &0.7800  &0.4820  &0.9182  &0.0177  \\
IRASim &0.6200  &\textbf{0.8427}  &0.9184  &0.7800  & 0.5620 & \textbf{0.9240} &0.0071  \\
\bottomrule
\end{tabular}
}
\end{table*}

\begin{table*}[h]
\centering
\caption{Video quality evaluation results across visual quality, motion quality and content consistency dimensions on the real-world robot.}
\label{tab:vq_real_part1}
\small
\setlength{\tabcolsep}{4pt}
\resizebox{\textwidth}{!}{
\begin{tabular}{lccccccccc}
\toprule
\multirow{2}{*}{\textbf{\large Models}} & \multicolumn{3}{c}{\textbf{\large Visual Quality}} & \multicolumn{3}{c}{\textbf{\large Motion Quality}} & \multicolumn{3}{c}{\textbf{\large Content Consistency}} \\ 
\cmidrule(lr){2-4} \cmidrule(lr){5-7} \cmidrule(lr){8-10}
 & \begin{tabular}[c]{@{}c@{}}Image \\ Quality\end{tabular} & \begin{tabular}[c]{@{}c@{}}Aesthetic \\ Quality\end{tabular} & \begin{tabular}[c]{@{}c@{}}JEPA \\ Similarity\end{tabular} & \begin{tabular}[c]{@{}c@{}}Dynamic \\ Degree\end{tabular} & \begin{tabular}[c]{@{}c@{}}Flow \\ Score\end{tabular} & \begin{tabular}[c]{@{}c@{}}Motion \\ Smoothness\end{tabular} & \begin{tabular}[c]{@{}c@{}}Subject \\ Consistency\end{tabular} & \begin{tabular}[c]{@{}c@{}}Backgrd. \\ Consistency\end{tabular} & \begin{tabular}[c]{@{}c@{}}Photom. \\ Consistency\end{tabular} \\ 
\midrule
GigaWorld-0 &0.5976  &0.4031  &0.1592  &0.4685  &0.2706  &0.6994  &0.7667  &0.9002  &0.3251  \\
Genie Envisioner & 0.3302 & 0.2428 &0.0041  &0.2440  &0.2031  &0.4390  &0.6691  &0.8569  &0.0776  \\
TesserAct &0.6045  &\textbf{0.4721}  &0.0236  &0.5263  &0.2032  &0.6883  &0.7932  &0.9070  &0.2760  \\
RoboMaster & 0.6050 &0.4258  &0.0249  &0.3556  &0.0938  & 0.5299 & \textbf{0.8137} &\textbf{0.9152}  &0.3593  \\
Vidar &0.6001 &0.4570  &0.0887  &0.1510  &0.0486  &0.5224  &0.4906  &0.5472  &\textbf{0.4149}  \\
Cosmos-Predict 2.5 (text) &\textbf{0.6903}  &0.4706  &0.0473  &0.3354  &0.1844  &0.6576  &0.7320  &0.8589  &0.3299  \\
Cosmos-Predict 2.5 (action) &0.5819  &0.3266  &0.1136  &0.2738  &0.1422  &0.5700  &0.4035  &0.4715  &0.1071  \\
WoW &0.6733  &0.4147  &0.1839  &0.4010  &0.1535  &0.6690  &0.7780&0.9113&0.2692  \\
CtrlWorld &0.6153  &0.3540  &\textbf{0.6621}  &\textbf{0.5649}  &\textbf{0.4963}  &\textbf{0.7456}  &0.7959  &0.9093  &0.0612  \\
Wan 2.2 &0.6267  &0.4552  &0.1047  &0.2979  &0.1935  &0.6434  &0.4964  &0.5871  &0.2339  \\
CogvideoX &0.6145  &0.3872  &0.4909  &0.3074  &0.1202  &0.5928  &0.7768  &0.8884  &0.2940  \\
IRASim &0.5175  &0.3327  &0.1056  &0.1845  & 0.0914 &0.5265  &0.4458  &0.5064  &0.1858  \\
\bottomrule
\end{tabular}
}
\end{table*}

\begin{table*}[htbp]
\centering
\caption{Video quality evaluation results across physics adherence, 3D accuracy and controllability dimensions on the real-world robot.}
\label{tab:vq_real_part2}
\small
\resizebox{\textwidth}{!}{
\begin{tabular}{lcccccccc}
\toprule
\multirow{2}{*}{\textbf{\large Models}} & \multicolumn{2}{c}{\textbf{\large Physics Adherence}} & \multicolumn{2}{c}{\textbf{\large 3D Accuracy}} & \multicolumn{3}{c}{\textbf{\large Controllability}} \\ 
\cmidrule(lr){2-3} \cmidrule(lr){4-5} \cmidrule(lr){6-8}
 & \begin{tabular}[c]{@{}c@{}}Interaction \\ Quality\end{tabular} & \begin{tabular}[c]{@{}c@{}}Trajectory \\ Accuracy\end{tabular} & \begin{tabular}[c]{@{}c@{}}Depth \\ Accuracy\end{tabular} & Perspectivity & \begin{tabular}[c]{@{}c@{}}Instruction \\ Following\end{tabular} & \begin{tabular}[c]{@{}c@{}}Semantic \\ Alignment\end{tabular} & \begin{tabular}[c]{@{}c@{}}Action Response \\Sensitivity\end{tabular} \\ 
\midrule
GigaWorld-0 &0.7450  &0.0815  &0.9258  &0.9238  &0.4012  &0.7816  &0.0277  \\
Genie Envisioner &0.2275  &0.0039  &0.6687  &0.4838  &0.2350  &0.8209  &0.0212  \\
TesserAct &0.6025  &0.0850  &0.8300  &0.8562  &0.5175  &0.8536  &0.0247  \\
RoboMaster &0.6812  &0.0585  &0.9164  &0.8863  &0.6987  &0.8779  &0.0343  \\
Vidar &0.7162  &0.1734  &0.9468  &0.9150  &0.7100  &0.9217  &0.0399  \\
Cosmos-Predict 2.5 (text) &\textbf{0.8600}  &0.0445  &0.5471  &0.9450  &0.9300  &0.8850  &\textbf{0.0574}  \\
Cosmos-Predict 2.5 (action) &0.7200  &0.2592  &0.9703  &0.8762  &0.8012  &0.8978  &0.0088  \\
WoW  &0.8100  &0.3603  &0.9448  &\textbf{0.9650}  &\textbf{0.9887}  &\textbf{0.9268}  &0.0399  \\
CtrlWorld &0.7387  &\textbf{0.6865}  &\textbf{0.9888}  &0.9275  &0.8838  & 0.8325 &0.0059  \\
Wan 2.2 &0.5662  &0.1164  &0.8381  &0.8637  &0.4237  &0.9092  &0.0548  \\
CogvideoX &0.7050  &0.3107  &0.9780  &0.9425  &0.8162  &0.9081  &0.0176  \\
IRASim &0.6837  &0.2621  &0.9545  &0.8475  &0.7963  &0.8873  &0.0140  \\
\bottomrule
\end{tabular}
}
\end{table*}

\end{document}